\documentclass[10pt,twocolumn,letterpaper]{article}
\pdfoutput=1
\usepackage{iccv}
\usepackage{times}
\usepackage{epsfig}
\usepackage{graphicx}
\usepackage{amsmath}
\usepackage{amssymb}
\usepackage{graphicx}
\usepackage{color}
\usepackage{bbding}
\usepackage{bm}
\usepackage{booktabs}
\usepackage{multirow}
\usepackage[noend]{algpseudocode}
\usepackage{algorithmicx,algorithm}
\usepackage{ifpdf}
 \ifpdf
 \else
 \fi

\usepackage[subfigure]{tocloft}
\usepackage{float}
\usepackage{graphicx}
\usepackage{subfigure}

\usepackage[breaklinks=true,bookmarks=false]{hyperref}

\iccvfinalcopy 

\def\iccvPaperID{6366} 

\ificcvfinal\fi

\begin{document}

\title{OSKDet: Towards Orientation-sensitive Keypoint Localization for Rotated Object Detection}

\author{Dongchen Lu\\
}

\maketitle

\begin{abstract}
Rotated object detection is a challenging issue of computer vision field. Loss of spatial information and confusion of parametric order have been the bottleneck for rotated detection accuracy. In this paper, we propose an orientation-sensitive keypoint based rotated detector OSKDet. We adopt a set of keypoints to characterize the target and predict the keypoint heatmap on ROI to form a rotated target. By proposing the orientation-sensitive heatmap, OSKDet could learn the shape and direction of rotated target implicitly and has stronger modeling capabilities for target representation, which improves the localization accuracy and acquires high quality detection results. To extract highly effective features at border areas, we design a rotation-aware deformable convolution module. Furthermore, we explore a new keypoint reorder algorithm and feature fusion module based on the angle distribution to eliminate the confusion of keypoint order. Experimental results on several public benchmarks show the state-of-the-art performance of OSKDet. Specifically, we achieve an AP of 77.81\% on DOTA, 89.91\% on HRSC2016, and 97.18\% on UCAS-AOD, respectively.

\end{abstract}

\section{Introduction}

With the success of deep convolutional neural networks (CNN), object detection has made an unprecedented breakthrough in recent years. General detection models\cite{fastrcnn,cornernet,yolo} regress the horizontal bounding box of objects. However, real scenes targets may have arbitrary directions, such as cars in drone cameras and texts in streetscape. Only determining the horizontal box is not enough to locate the target accurately. Rotated object detection has a wide range of applications, but still faces great challenges.


\begin{figure}[htbp]
  \centering 
  \subfigure[]{ 
    \label{fig1_1} 
    \includegraphics[width=0.77in]{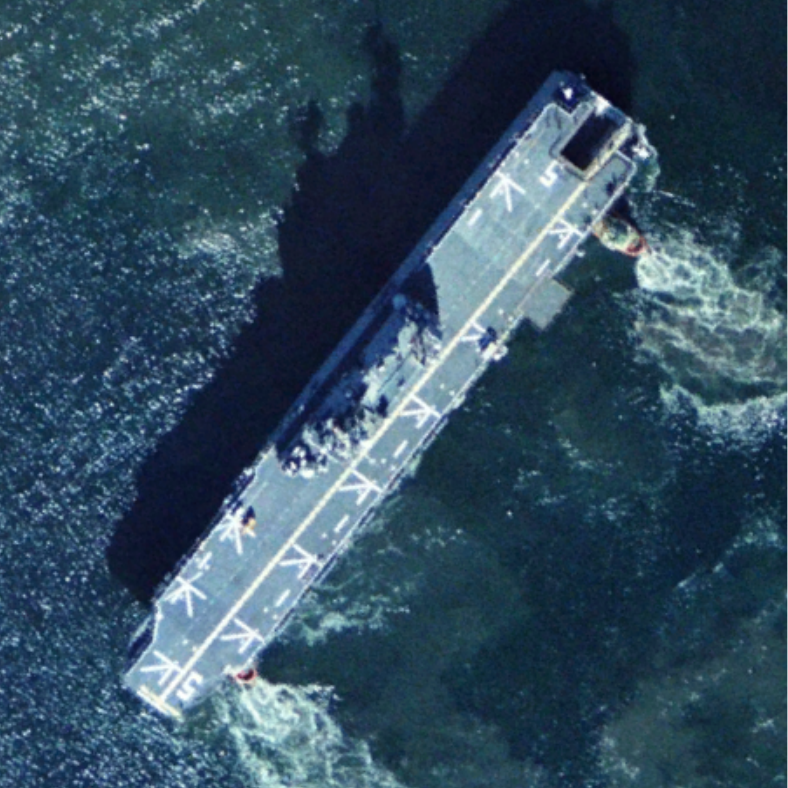} 
  }\hspace{-1ex}
  \subfigure[]{ 
    \label{fig1_2} 
    \includegraphics[width=0.77in]{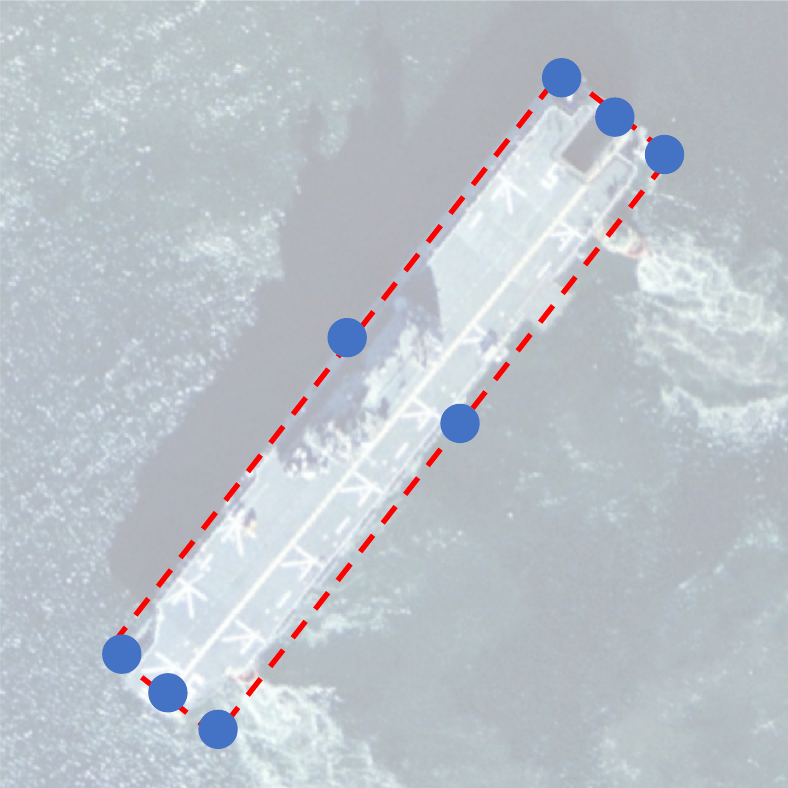} 
  }\hspace{-1ex}
  \subfigure[]{ 
    \label{fig1_3} 
    \includegraphics[width=0.77in]{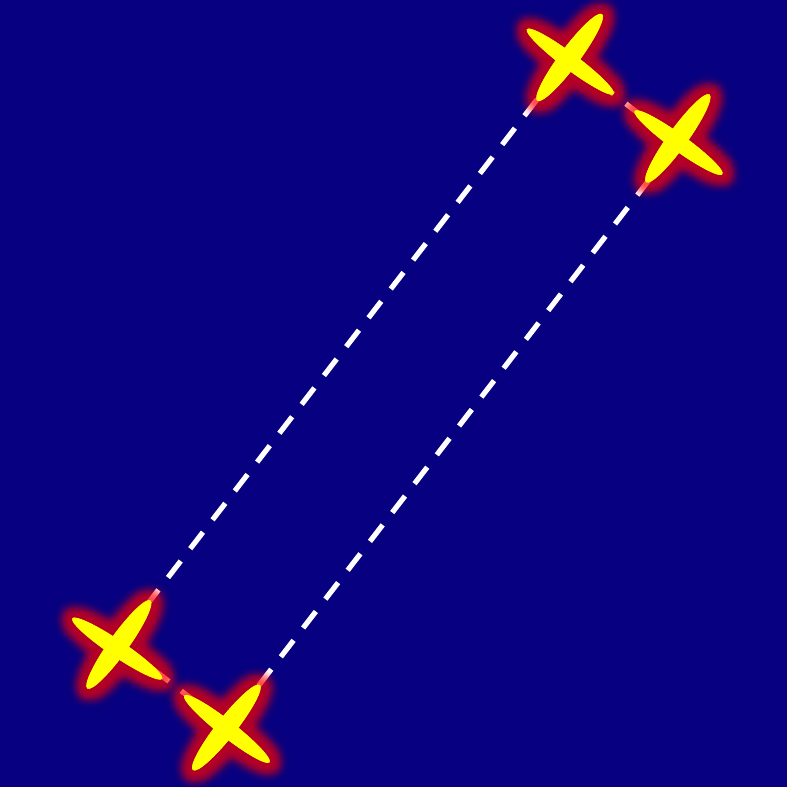} 
  }\hspace{-1ex}
  \subfigure[]{ 
    \label{fig1_4} 
    \includegraphics[width=0.77in]{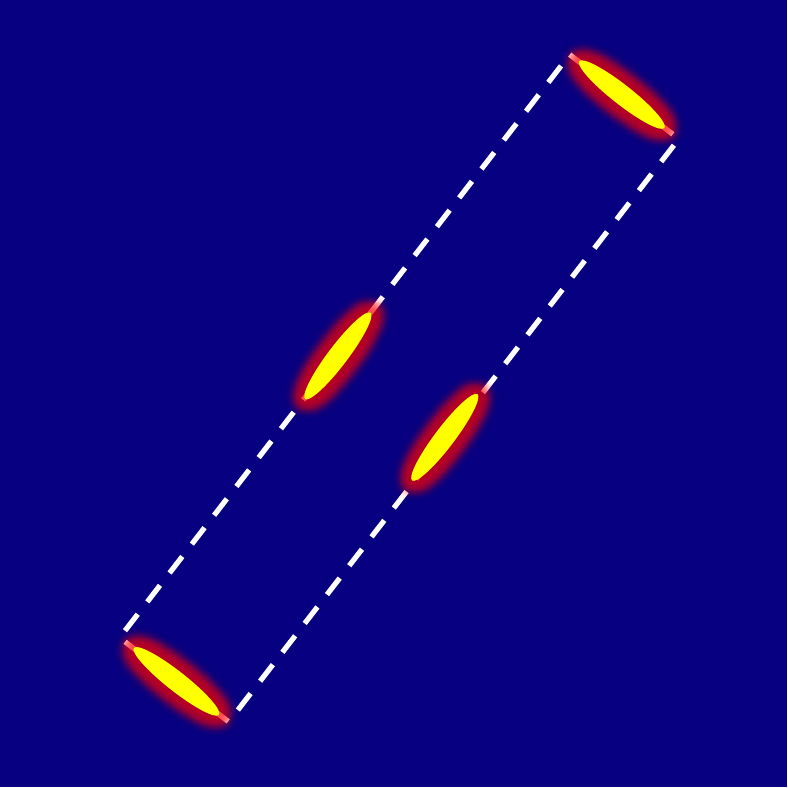} 
  }\vspace{-1ex}
  \subfigure[]{ 
    \label{fig1_5} 
    \includegraphics[width=3.2in]{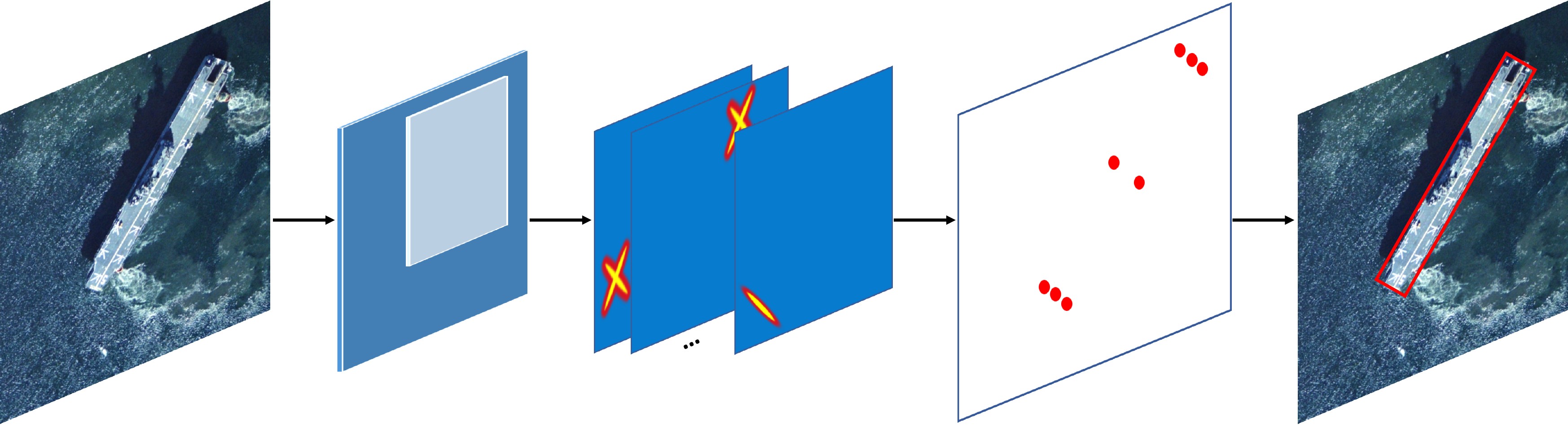} 
  } 
   
  \caption{(a) a rotated target represented by 8 keypoints in (b). (c) and (d) display the proposed orientation-sensitive heatmap, we encode the keypoint to cross-star shape in corner and straight shape in edge areas, which represents the outline of target more accurately. (e) OSKDet: keypoint based detector. OSKDet generates the orientation-sensitive keypoint heatmap through fully convolution structure, which has a strong modeling ability of spatial representation and achieves high quality detections.} 
  \label{fig1} 
\end{figure}

Recently, CNN based rotated detection has made a considerable progress. Some works \cite{piou,roitransformer,rrpn,scrdet,r2pn} use the angle definition (coordinates of the central point, width, height, and rotated angle) to represent the rotated target. Other works \cite{ienet,sbd,rsdet,glide_vertex,polardet} use vertex definition (coordinates of four vertices) to describe the rotated quadrilateral. Mainstream models mainly face the following issues: i) loss of spatial information, as traditional regression based detection models utilize spatial information insufficiently, and will lose this relative location information. For rotated targets with more random positions and directions, the missing of spatial information weakens the capability of localization; ii) confusion of parametric order, which is caused by the parametric exchangeability and definition periodicity of rotated targets, and manual labeling errors exacerbate this problem. Similar samples with different optimization direction will cause learning confusion. In fact, these issues can largely affect the detection performance especially at the localization accuracy.

In this paper, we propose a keypoint based rotated detector OSKDet. We encode a set of keypoints to represent rotated target. To avoid the loss of spatial information, we drop the fully connected layer and adopt the fully convolution block to generate keypoint heatmap on ROI. Considering that rotated target has more obvious features at the vertex and edge areas, we design an orientation-sensitive heatmap, as shown in Fig \ref{fig1_3} and Fig \ref{fig1_4}, which could better match the target shape, and the network could learn the orientation and shape of rotated target implicitly. OSKDet has stronger modeling capabilities in spatial representation and transformation. Furthermore, we propose a rotation-aware deformable convolution module to extract the target border feature more effectively. We also explore a keypoint reorder algorithm based on the dataset angle distribution, which could eliminate the keypoint order confusion problem to the greatest extent.

The proposed three modules can notably improve rotated detection accuracy. Extensive experiments on public benchmarks, including aerial dataset DOTA\cite{DOTA}, HRSC2016\cite{hrsc2016}, UCAS-AOD\cite{ucas-aod}, scene text dataset ICDAR2015\cite{ICDAR2015} and ICDAR2017MLT\cite{icdar2017}, show the superiority of OSKDet.

Our main contributions can be summarized as follows:

1) We propose a keypoint based rotated detector OSKDet that predicts the keypoint heatmap on the ROI. We design an orientation-sensitive heatmap, which could learn the shape and direction of rotated target, and plays a significant role in improving the localization accuracy.

2) We propose a rotation-aware deformable convolution module, which could extract the features of target border regions more effectively.

3) We explore a new angle distribution based keypoint reorder algorithm to eliminate keypoint order confusion. And on this basis, a new local and global feature fusion module is proposed.



\section{Related work}
\subsection{Horizontal object detection}

Detection models with deep neural networks achieved superior performance on the public datasets COCO\cite{coco} and VOC\cite{voc} recently. According to whether there is a series of candidate anchor box, detection models can be divided into anchor based and anchor free methods. While according to the final localization mode, existing models have regression and heatmap methods. Most anchor based detectors localize targets by regression mode. Faster RCNN\cite{fasterrcnn}, FPN\cite{fpn}, Cascade RCNN\cite{cascade} etc. use fully connected layers to predict the deviation between anchor box and target. YOLO v2\cite{yolov2} and YOLO v3\cite{yolov3} adopt fully convolution structure and regress target center point ratio on each grid cell. Recently some anchor free methods detect targets mainly by generating heatmap of keypoints. CenterNet\cite{centernet}, CornerNet\cite{cornernet}, ExtremeNet\cite{extremenet} etc. predict the probability of center or corner point in the whole image, and then group the points to form a box. Grid RCNN\cite{gridrcnn} predicts the keypoint heatmap on ROI and acquires a higher localization accuracy compared to Faster RCNN\cite{fasterrcnn}.

\subsection{Rotated object detection}
Similar to the horizontal object detection, mainstream rotated detection algorithms can be divided into anchor based and anchor free models, in which anchor based models have horizontal and rotated anchor box. According to the definition of rotated target, there are two types of prediction mode: angle representation and vertex representation. Angle definition uses the rotated angle($\theta$) of the longer border around the horizontal axis as the fifth parameter, with other parameters constitute the expression paradigm of rotated target ($x,y,w,h,\theta$). Vertex representation indicates rotated target by marking four vertices of the quadrilateral ($x_{0},y_{0},x_{1},y_{1},x_{2},y_{2},x_{3},y_{3}$). Compared to angle notation, vertex notation can represent any shape quadrilateral, and as mentioned in \cite{rsdet}, the four vertices regression has natural consistency, which means it is easier to optimize.

\textbf{Angle based detector.}
 DRBox\cite{DRBOX} and R2PN\cite{r2pn} introduce rotated anchor box based on rotated RPN. 
RRPN\cite{rrpn} proposes rotated ROI pooling to extract feature more effectively. R2CNN\cite{r2cnn} regresses two adjacent vertices and another side length of rotated target. ROI transformer\cite{roitransformer} transforms horizontal proposals to rotated ones through fully connection learning in RPN stage. EAST\cite{EAST} regresses the distance between each pixel and four sides of rotated box. SCRDet\cite{scrdet} highlights the target features through attention module, and proposes IOU smooth L1 loss to smooth the boundary loss jump. CSL\cite{csl} transforms angle prediction from regression to classification, and proposes circular label smooth. PIOU\cite{piou} adopts pixel counting method to calculate polygon IOU and makes it approximately derivable. 

\textbf{Vertex based detector.}
 Textboxes++\cite{textboxes++} proposes irregular long convolution kernel to adapt to targets with large aspect ratio. 
Gliding Vertex\cite{glide_vertex} regresses the ratio of the four vertices relative to the four points of horizontal box, and proposes an obliquity factor distinguishing nearly horizontal and other rotated objects. PolarDet\cite{polardet} adopts several vertices angles and length ratio around the center point in polar coordinates to generate rotated box. SBD\cite{sbd} proposes sequential-free box discretization parameterizing rotated boxes into key edges to eliminate the sensitive of label sequence. RSDet\cite{rsdet} proposes modulated loss by swapping point set sequence to obtain the best optimization direction. 

\section{Proposed methods}

\textbf{Overview.} In this section, we first present two main issues that hinder the rotated detection accuracy, and then introduce our method.

\textbf{Loss of spatial information.} Mainstream rotated detector adopt fully connected layers and regression method to localize target. As \cite{gap} proposed the flattening operation of fully connected layers will lose  spatial context information. While the convolution module has the ability to locate target and maintain  these relative information, and there is a pixel mapping relationship between the image and feature map. Compared to horizontal detection, rotated target has more complex spatial location diversity and transform feature, especially in the border regions. The spatial information plays a considerable role in localization accuracy.
\vspace{-1ex}
\begin{figure}[htbp]
    \label{fig2} 
    \includegraphics[width=\linewidth]{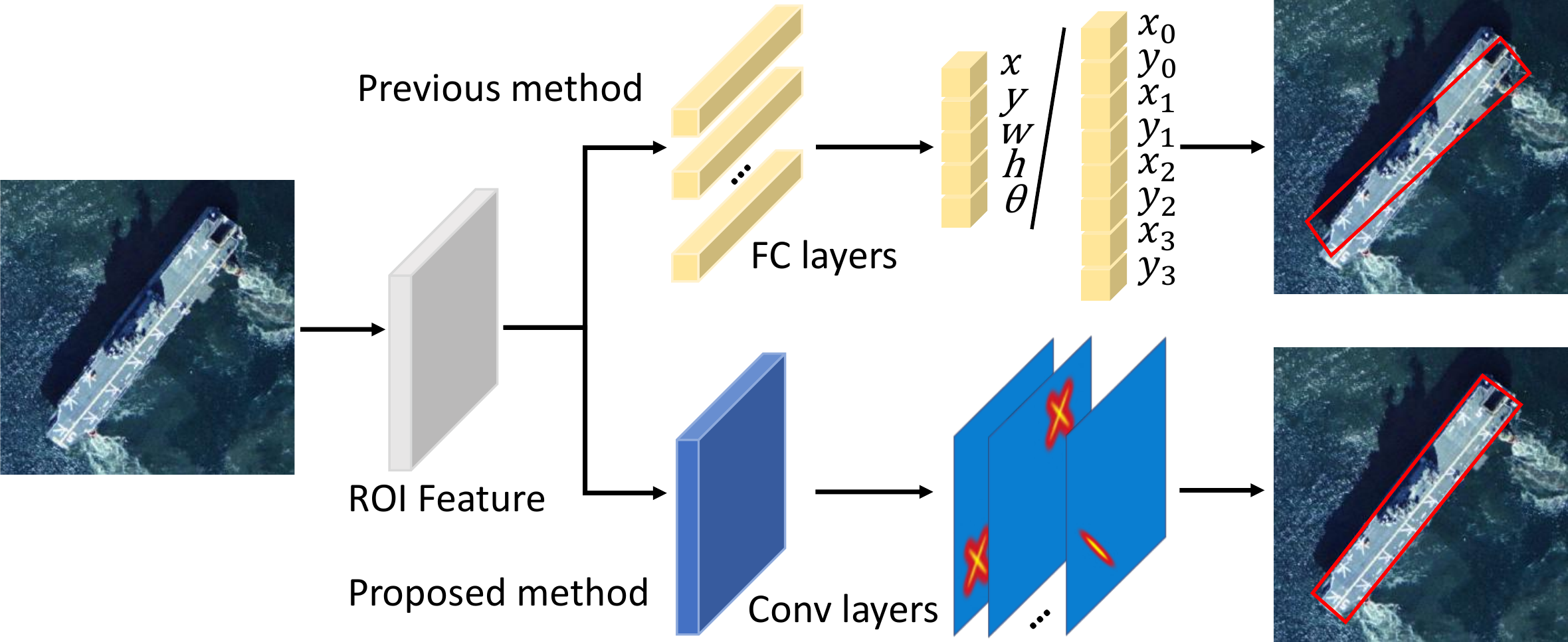}
    \vspace{0.1ex}
    \caption{Previous method: previous works regress rotated angle or coordinates of vertex through fully connected layers. Proposed method: OSKDet predicts the keypoint by generating heatmap through full convolution block, which can preserve the spatial information loss and has stronger spatial generalization ability.}     
\end{figure}
\vspace{-3ex}

\begin{figure}[htbp]
    \begin{center}
      \subfigure[]{ 
        \label{fig3_1} 
        \includegraphics[width=0.8in]{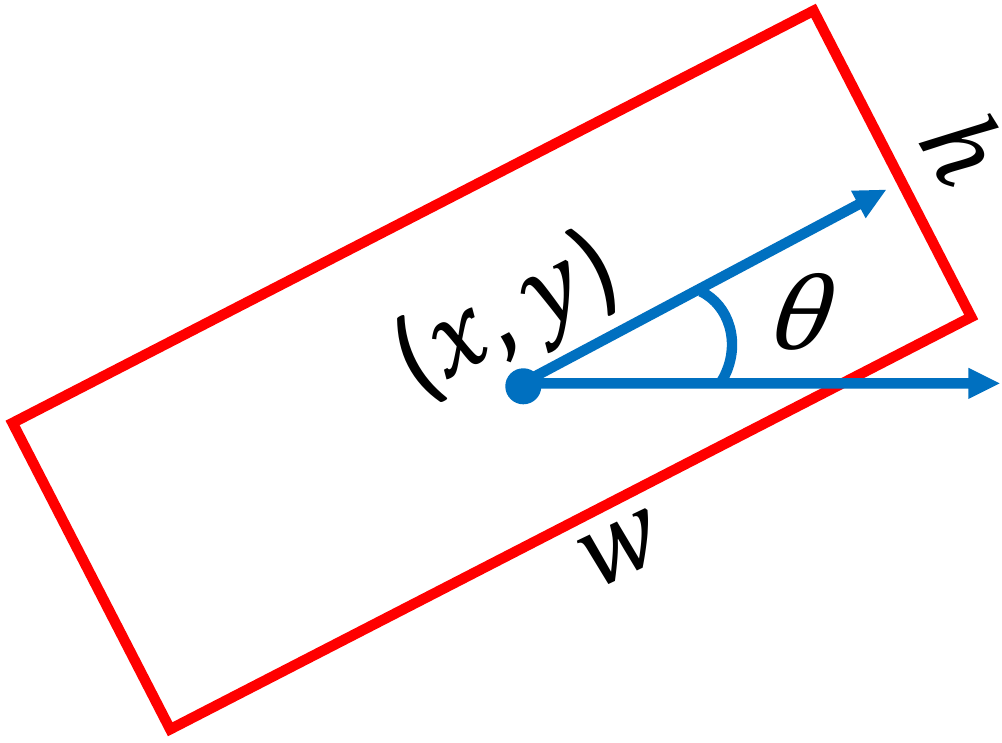} 
      }\hspace{-1.2ex}
      \subfigure[]{ 
        \label{fig3_2} 
        \includegraphics[width=0.8in]{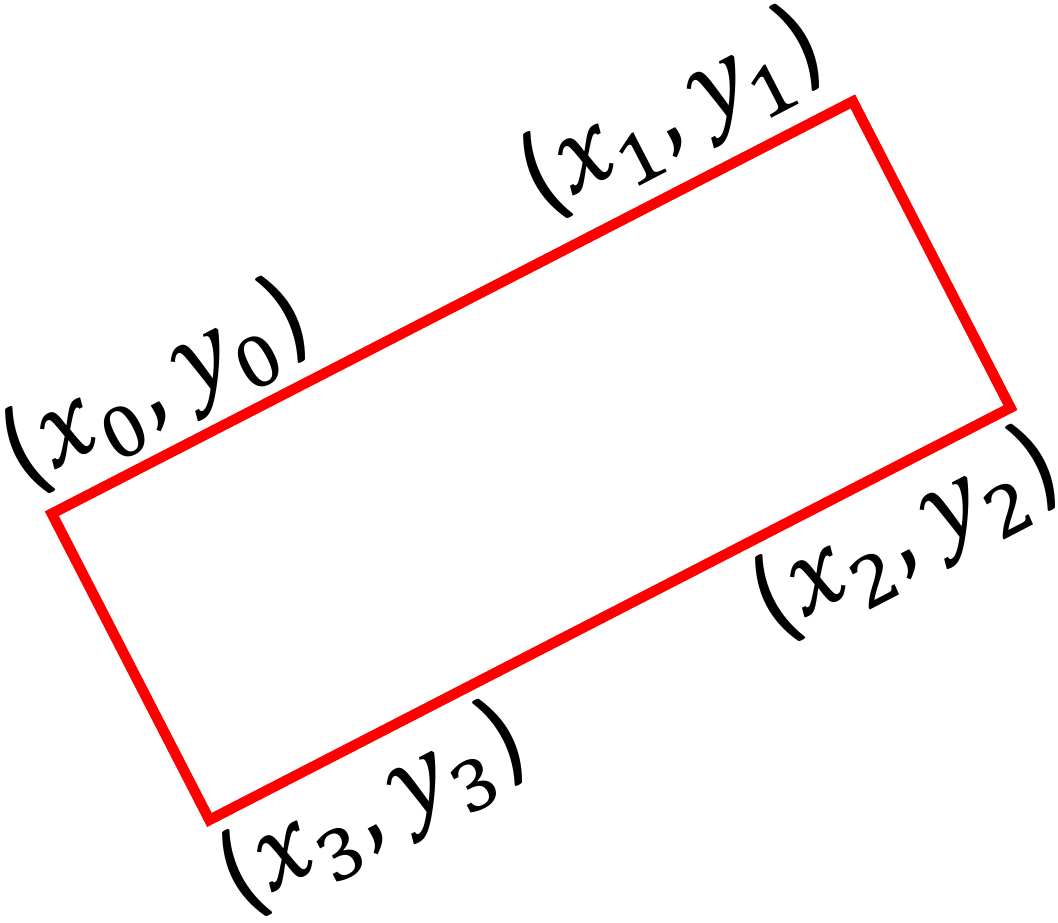} 
      }\hspace{-1.6ex}
      \subfigure[]{ 
        \label{fig3_3} 
        \includegraphics[width=0.77in]{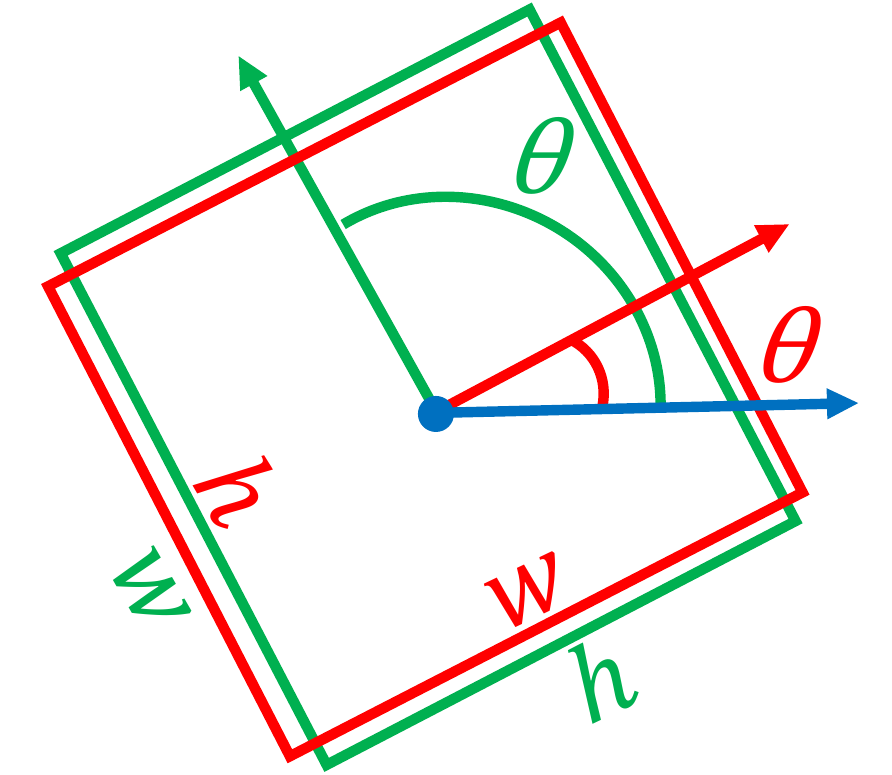} 
      }\hspace{-1.8ex}
      \subfigure[]{ 
        \label{fig3_4} 
        \includegraphics[width=0.8in]{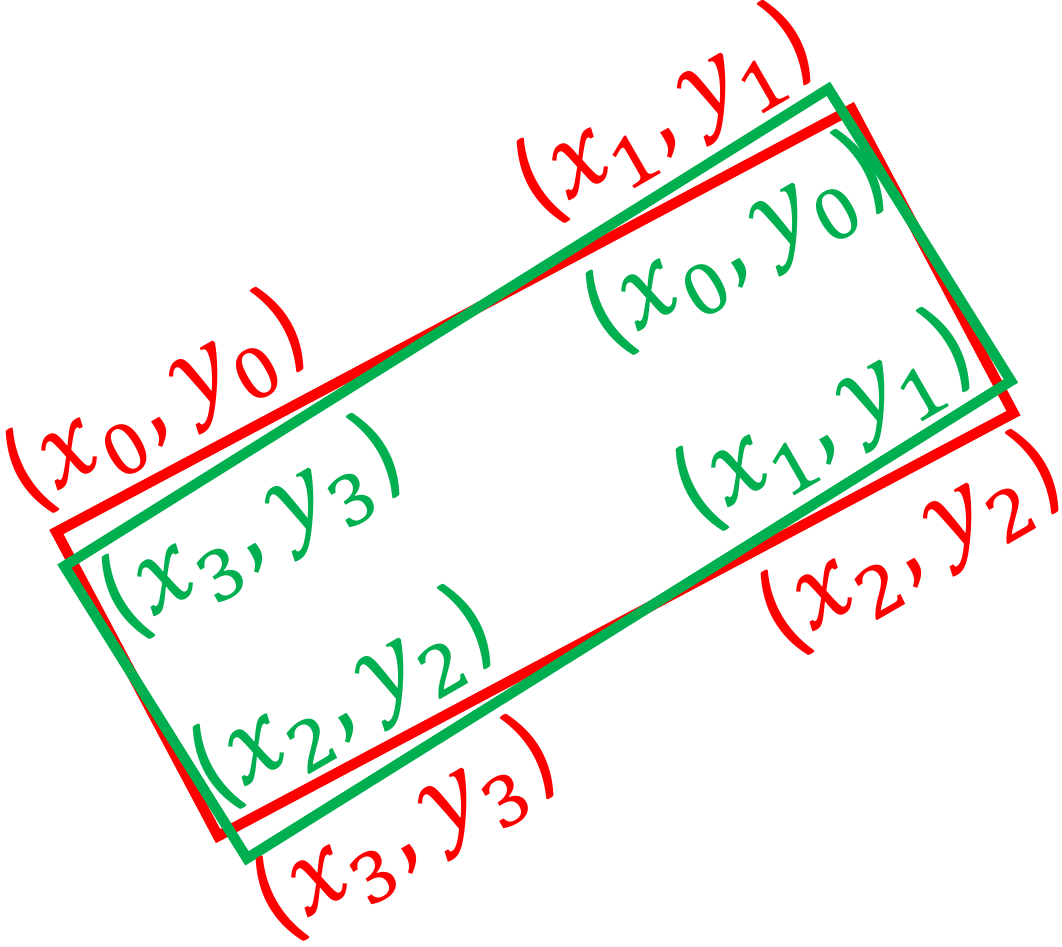} 
      }
    \end{center}
\vspace{-1.5ex}
      \caption{Different rotation representation: (a) angle notation. (b) vertex notation. (c) and (d) show parameters confusion. In (c), the red and green targets are very similar, while different length sequence causes width and height exchange, and their angle has $\pi$/2 gap. In (d), the red and green target has a slight rotation deviation,
       which leads to the vertex order one bit clockwise move.} 
      \label{fig3} 
    \end{figure}

\textbf{Confusion of parametric order.} Both two definition methods have learning confusion problems, especially in the definition boundaries. For the angle representation, as shown in Fig \ref{fig3_3}, when the target is near a standard square, a little borderline length variety may lead to the exchange of width and height, and there is a $\pi$/2 jump in angle. Similar phenomena appear in the vertex representation, as illustrated in Fig \ref{fig3_4}, a slight spin may cause the vertex order transform. In the training process, two similar samples with quite different labels will lead to completely different optimization directions. 
  

\vspace{-1ex}
\vspace{-1ex} 

The proposed method is based on the horizontal anchor box. 
Inspired by Grid RCNN\cite{gridrcnn}, we predict the keypoint heatmap on horizontal ROI. Specifically, we adopt four vertices and four edge midpoints to represent a complete rotated target. The network architecture is shown in Fig \ref{fig4}. After deep feature extraction and RPN filtering, we feed the multi-path fused ROI feature into the keypoint head, and get fine-grained keypoint heatmap through twice deconvolution. We design a new orientation-sensitive heatmap and a rotation-aware deformable convolution module. OSKDet has two optional branch outputs. Through the red branch, OSKDet outputs the orientation-sensitive heatmap. Through the blue branch, the network can obtain more accurate localization through the rotation-aware deformable convolution. Furthermore, we explore a minimum confusion keypoint reorder method as a preprocess module in the training stage. We will introduce each module in detail next.

\subsection{Orientation-sensitive heatmap}

\cite{maskrcnn,cornernet,gridrcnn} propose different heatmap generation methods. Gaussian distribution heatmap adopted by\cite{centernet,cornernet,extremenet} is most widely used. Different encoding methods affect the final detection accuracy. In this section, we propose a novel orientation-sensitive heatmap (OSH), which is different from previous work. Next, we will introduce the generation mode of the new heatmap and the comparison with the standard gaussian heatmap (SGH).
\vspace{-1ex}
\begin{figure}[htbp]
  \begin{center}
    \subfigure[]{ 
      \label{fig5_1} 
      \includegraphics[width=0.776in]{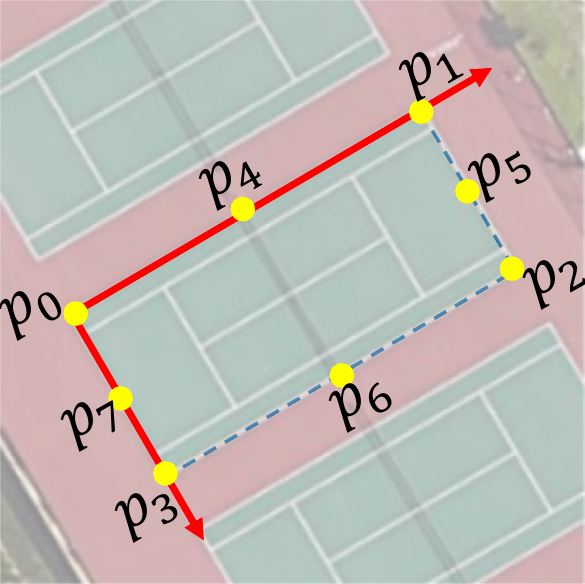}
    }\hspace{-1.05ex}
    \subfigure[]{ 
      \label{fig5_2} 
      \includegraphics[width=0.776in]{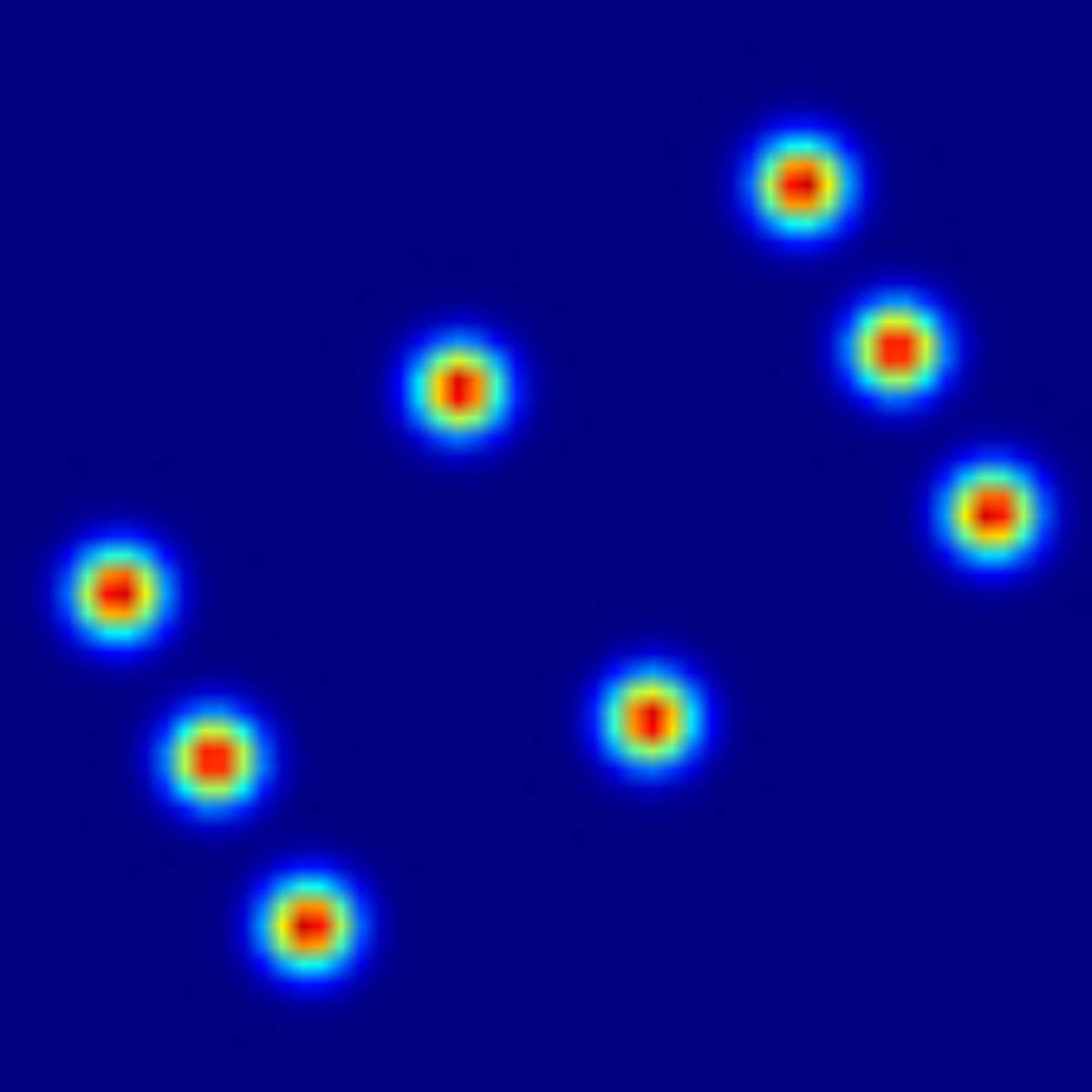}
    }\hspace{-1.05ex}
    \subfigure[]{ 
      \label{fig5_3} 
      \includegraphics[width=0.776in]{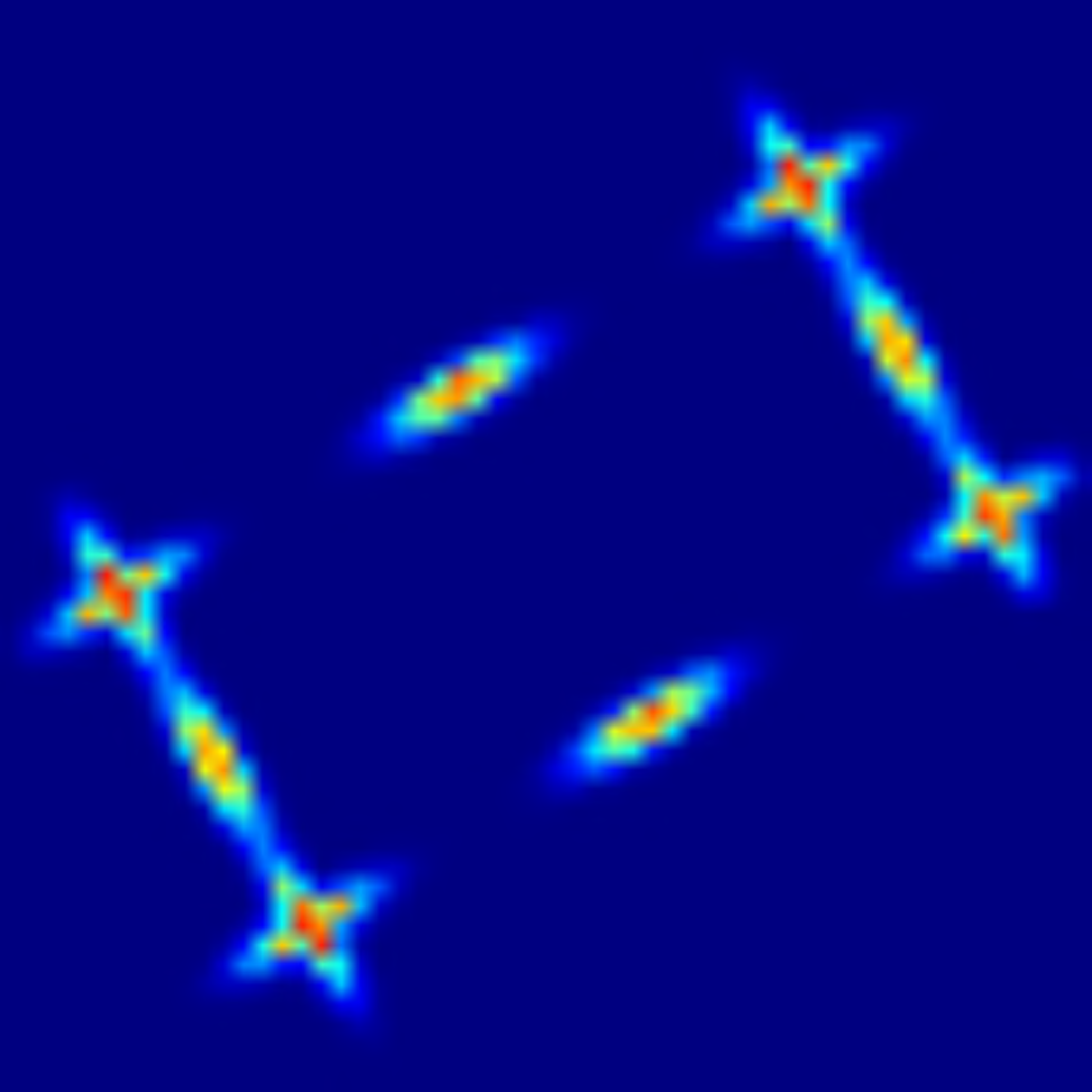}
    }\hspace{-1.05ex}
    \subfigure[]{ 
      \label{fig5_4} 
      \includegraphics[width=0.776in]{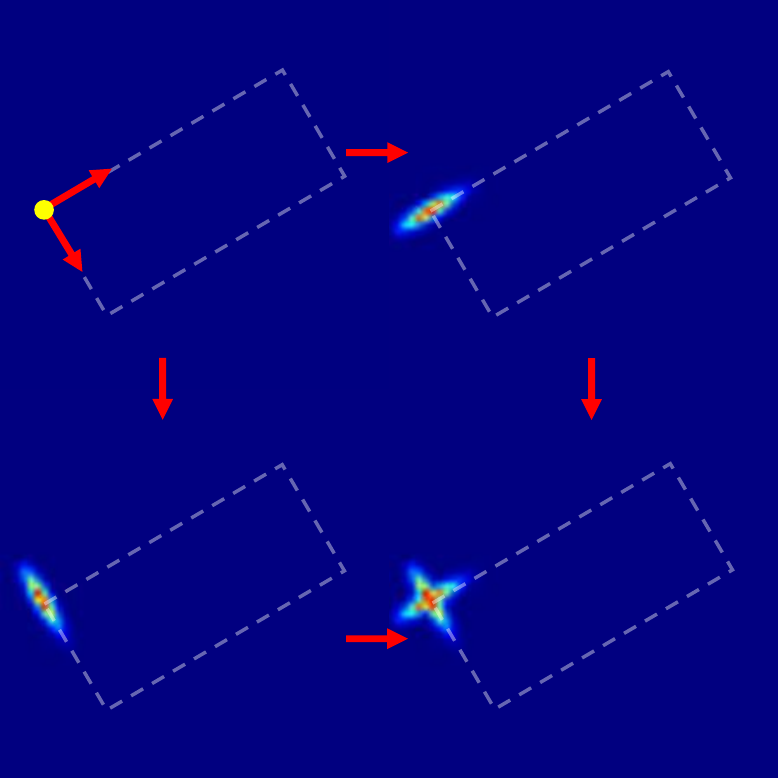}
    }
  \end{center}
  \vspace{-1ex}
    \caption{(a) a set of 8 ordered points representing a rotated target, including 4 vertices $p_{0-3}$ and 4 midpoints $p_{4-7}$. (b) and (c) are the comparison of SGH and proposed OSH. Compared to the SGH with same variance in $x$ and $y$ axis, the OSH assigns different weights according to the importance of the area, which more accurately reflects the contours of vertex and edge. (d) displays the generation of OSH for the vertex $p_{0}$.} 
    \label{fig5} 
  \end{figure}
\vspace{-1ex}

For the $p_{0}$ point in Fig \ref{fig5_1}, compared with other regions, the intersection area of $p_{0}p_{1}$ and $p_{0}p_{3}$ have more obvious color and texture transformation, and the edge between two vertices has similar feature. We believe that the junction zone information is more important, and hope that the network output has higher response values in these areas. The proposed OSH encodes different response values according to the importance of spatial region. In Fig \ref{fig5_1}, we establish new coordinate axis in the direction of $p_{0}p_{1}$ and $p_{0}p_{3}$ respectively, in which direction the gaussian distribution has greater variance, as shown in Fig \ref{fig5_3}. OSH has stronger spatial representation and transformation ability.

\begin{figure*}[htbp]
  \begin{center}
  \includegraphics[width=6.87in]{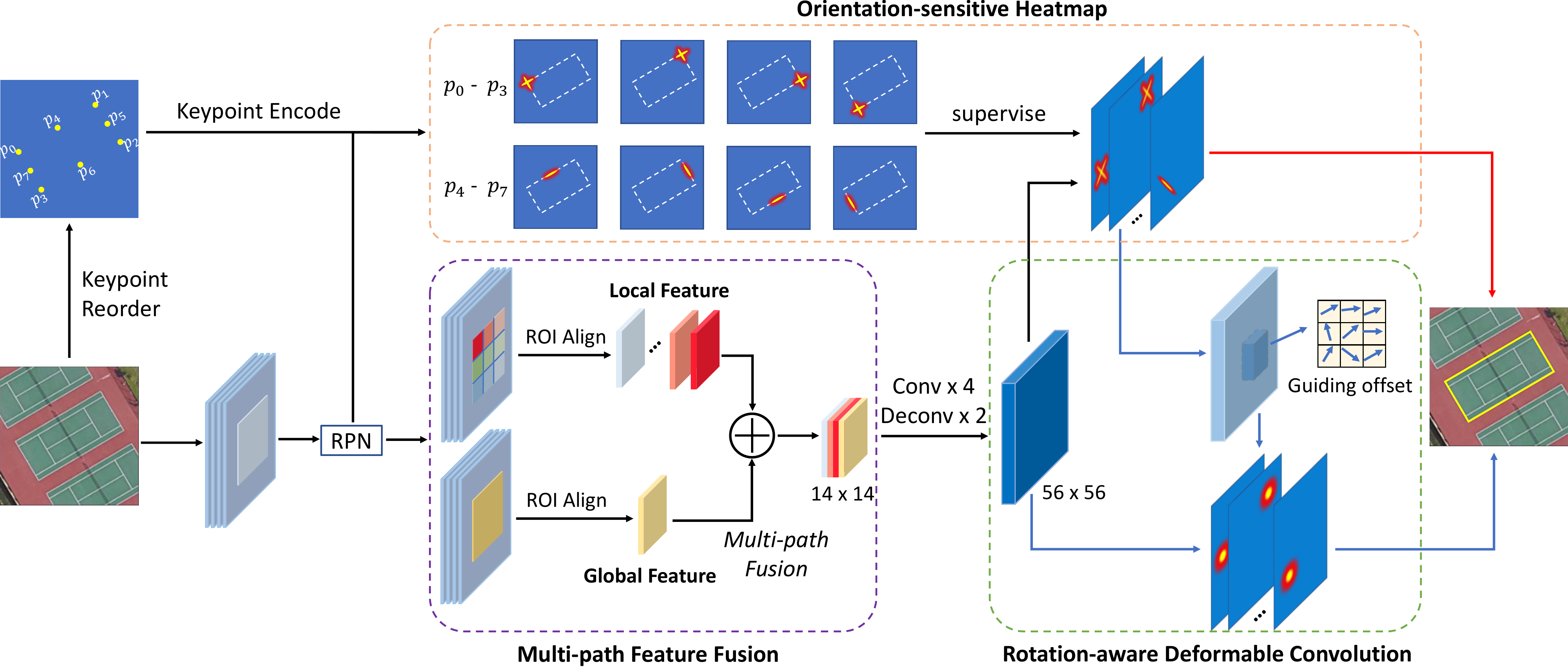}
  \end{center}
  \vspace{-1ex}
      \caption{Architecture of the proposed OSKDet. OSKDet contains a keypoint reorder preprocess module (KR), a multi-path feature fusion module (MFF), an orientation-sensitive heatmap modeling (OSH) and a rotation-aware deformable convolution module (RCN). OSKDet has two optional output branches. After the MFF fusion, the red branch generates the OSH format heatmap, by which we get the keypoint representation of the target spatial modeling. The blue branch take the OSH as the RCN coordinate offset learning input, and get a more accurate localization through RCN. We get final detections after decoding either heatmap.}
  \label{fig4}
  \end{figure*}


After getting the proposal ROI from RPN, we map the ground truth keypoints to the ROI space. 
In order to include as many keypoints as possible, we adopt the ROI expansion idea of Grid RCNN\cite{gridrcnn}, and expand the ROI from the center to $(1+r)$ times size ($r$ is set to 0.25). Assuming that the original ROI region is $\left(x, y, w, h\right)$, where $\left(x, y\right)$ is the left-top point of the ROI, we obtain the expanded ROI  $\left(x^{\prime}, y^{\prime}, w^{\prime}, h^{\prime}\right)$ through Eq \ref{eq1}.

\vspace{-1ex}
\begin{equation}\label{eq1}
\begin{array}{cc}
x^{\prime}=x-\frac{r}{2} w, \quad y^{\prime}=y-\frac{r}{2} h\vspace{1ex} \\
w^{\prime}=(1+r) w, \quad h^{\prime}=(1+r) h
\end{array} 
\end{equation}
\vspace{-1ex}

Define the ground truth keypoints $P=\left\{p_{k}\right\}_{k \in[0, K-1]}$, $p_{k}=\left(\hat{x_{k}}, \hat{y_{k}}\right)$ and mapped keypoints $Q=\left\{q_{k}\right\}_{k \in[0, K-1]}$, $q_{k}=\left(x_{k}, y_{k}, \theta_{k}\right)$. we adopt a total of $K = 8$ keypoints. After twice deconvolution, We will get a heatmap of $M \times M$ size (default $M = 56$). Through Eq \ref{eq2}, we calculate the mapped coordinates on the heatmap. For the mapped points beyond ROI boundary, the generated heatmap is all 0.

\vspace{-1.5ex}
\begin{equation}\label{eq2}
x_{k}=\left(\hat{x_{k}}-x^{\prime}\right) \frac{M}{w^{\prime}}, \quad y_{k}=\left(\hat{y_{k}}-y^{\prime}\right) \frac{M}{h^{\prime}}
\end{equation}
\vspace{-2ex}

For the vertex $q_{0-3}$, we calculate the rotated angle $\theta$ in the direction of two adjacent sides (e.g. for $q_{0}$, we calculate the angle of $q_{0}q_{1}$ and $q_{0}q_{3}$ directions).
\begin{equation}\label{eq3}
\begin{array}{l}
\theta_{k}=\left[\operatorname{argtan}\left(\frac{y_{(k+1) \% 4}-y_{k}}{x_{(k+1) \% 4}-x_{k}}\right),   \operatorname{argtan}\left(\frac{y_{(k-1) \% 4}-y_{k}}{x_{(k-1) \% 4}-x_{k}}\right)\right]
\end{array}
\end{equation}

For the midpoint $q_{4-7}$, we only calculate the rotated angle formed by the vertices on both sides (e.g. for $q_{4}$, we calculate the angle of $q_{0}q_{1}$ direction).

\vspace{-1.8ex}
\begin{equation}\label{eq4}
\theta_{k}=\operatorname{argtan}\left(\frac{y_{(k-3) \% 4}-y_{k-4}}{x_{(k-3) \% 4}-x_{k-4}}\right)
\end{equation}

Suppose the mean and covariance matrix of standard gaussian distribution be $\bm{\mu_{k}}=\left(x_{k}, y_{k}\right)^{T}$ , $\bm{\Sigma}=$ $\textstyle\left[\begin{array}{cc}
\sigma_{11} & 0 \\
0 & \sigma_{22}
\end{array}\right]$. 
The rotation matrix of cartesian coordinate about the origin is $\bm{{R_{k}}}=$ $\left[\begin{array}{cc}
\cos \theta_{k} & -\sin \theta_{k} \\
\sin \theta_{k} & \cos \theta_{k}
\end{array}\right]$. 
The covariance matrix after rotation transformation is

\vspace{-1.7ex}
\begin{equation}\label{eq5}
\bm{\Sigma_{k}}=\bm{R_{k}^{T} \Sigma R_{k}}
\end{equation}
\vspace{-2.4ex}

The final OSH is generated by Eq \ref{eq6}. For the vertex $q_{0-3}$, we generate heatmap in two directions. Fig \ref{fig5_4} displays the OSH generation process.
\vspace{-0.5ex}
\begin{small}
\begin{equation}\label{eq6}
\bm{H(g)_{k}}=\frac{1}{\sqrt{2 \pi|\bm{\Sigma_{k}}|}} \exp \left[-\frac{1}{2}\left(\bm{g}-\bm{\mu_{k}}\right)^{T} \bm{\Sigma_{k}}^{-1}\left(\bm{g}-\bm{\mu_{k}}\right)\right]
\end{equation}
\end{small}
\vspace{-1.8ex}


We adopt the binary cross entropy loss function to optimize the regression error of the heatmap, as illustrated by Eq \ref{eq7}. The $\delta(\cdot)$ is sigmoid function.

\vspace{-2.2ex}
\begin{equation}\label{eq7}
\begin{array}{c}
Loss=\frac{1}{K \times M \times M} \sum\limits_{k=0}^{K-1} \sum\limits_{i=0}^{M-1} \sum\limits_{j=0}^{M-1} -\left( \hat{h_{k i j}}\right) \log \left(\delta\left(h_{k i j}\right)\right) \vspace{1ex}\\
-\left(1-\hat{h_{k i j}}\right) \log \left(1-\delta\left(h_{k i j}\right)\right) \vspace{1ex}\\

\end{array}
\end{equation}
\vspace{-2.3ex}
\vspace{-1ex}
\subsection{Rotation-aware deformable convolution}

\begin{figure}[htbp]
\begin{center}
\includegraphics[width=2.8in]{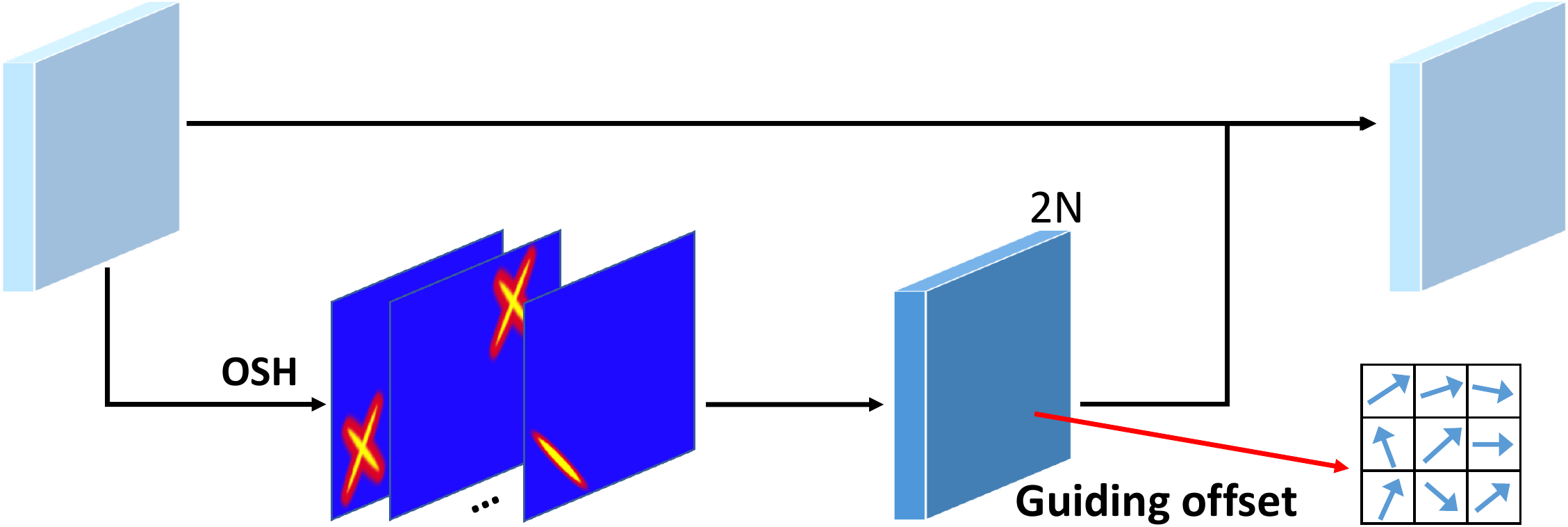}
\end{center}
\vspace{-0.8ex}
  \caption{Rotation-aware deformable convolution module: RCN takes OSH format heatmap as coordinate offset learning input, and generates more accurate sampling points for feature extraction.}
  \label{fig6}
\end{figure}

The vertex and edge regions of rotated target have more valuable features. 
We hope that the model can capture more effective features at these important areas to form a kind of spatial attention, which needs the convolution kernel to have the feature extraction capabilities for irregular shapes. Inspired by deformable convolution\cite{DCN} and cross-star deformable convolution\cite{centripetalnet}, we propose a new rotation-aware convolution (RCN), as shown in Fig \ref{fig6}. The orientation-sensitive heatmap in previous section acts as an ``attention module" to guide the generation of coordinate offset. The OSH accurately represents rotated target in the vertex and edge regions. By learning the geometric structure of OSH, the network will be more attentive to these valuable areas.

\vspace{-1ex}
\begin{figure}[htbp]
\begin{center}
  \subfigure[CNN]{ 
    \label{fig7_1} 
    \includegraphics[height=0.65in]{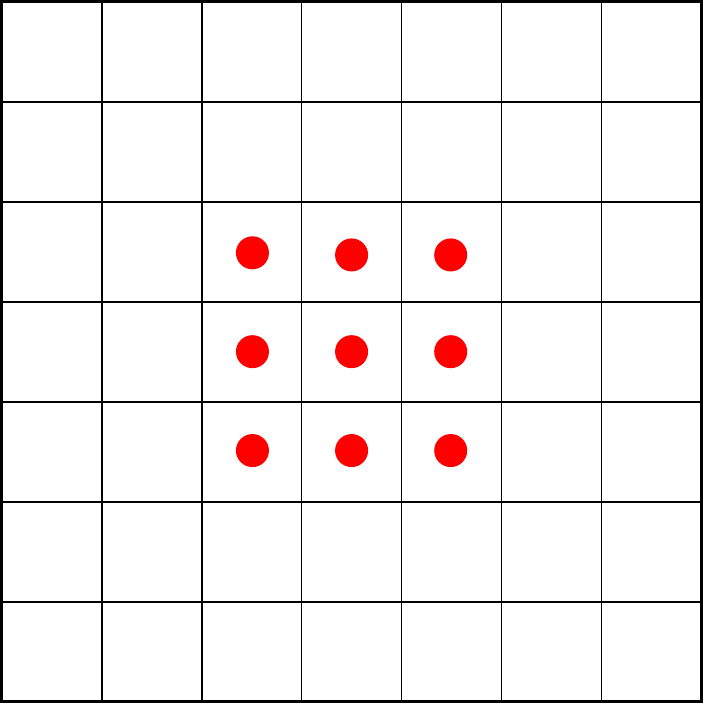} 
  }
  \subfigure[DCN]{ 
    \label{fig7_2} 
    \includegraphics[height=0.65in]{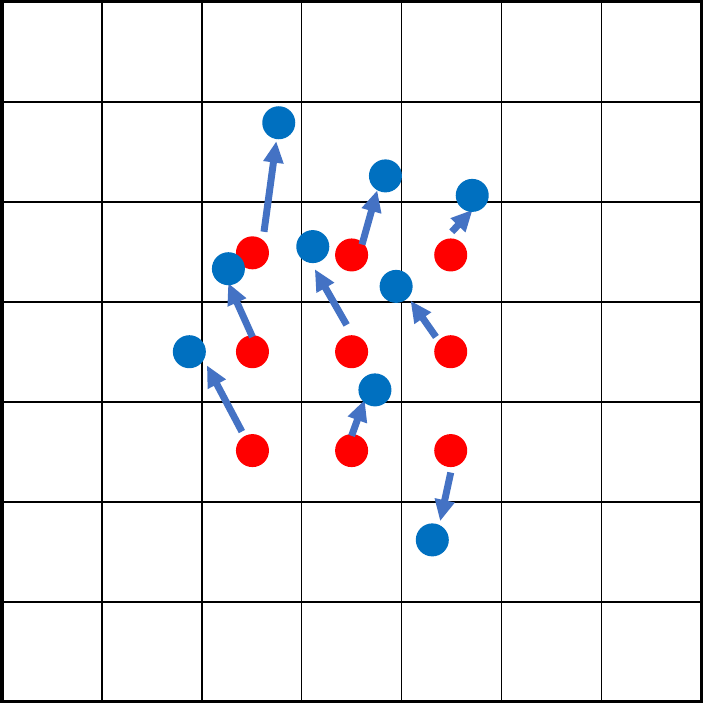} 
  } 
  \subfigure[RCN-vertex]{ 
    \label{fig7_3} 
    \includegraphics[height=0.65in]{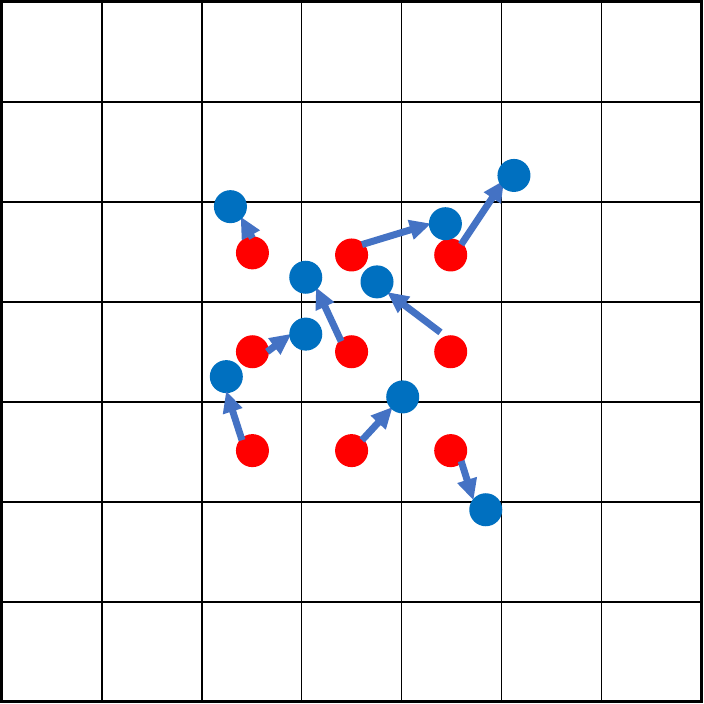} 
  }
  \subfigure[RCN-edge]{ 
    \label{fig7_4} 
    \includegraphics[height=0.65in]{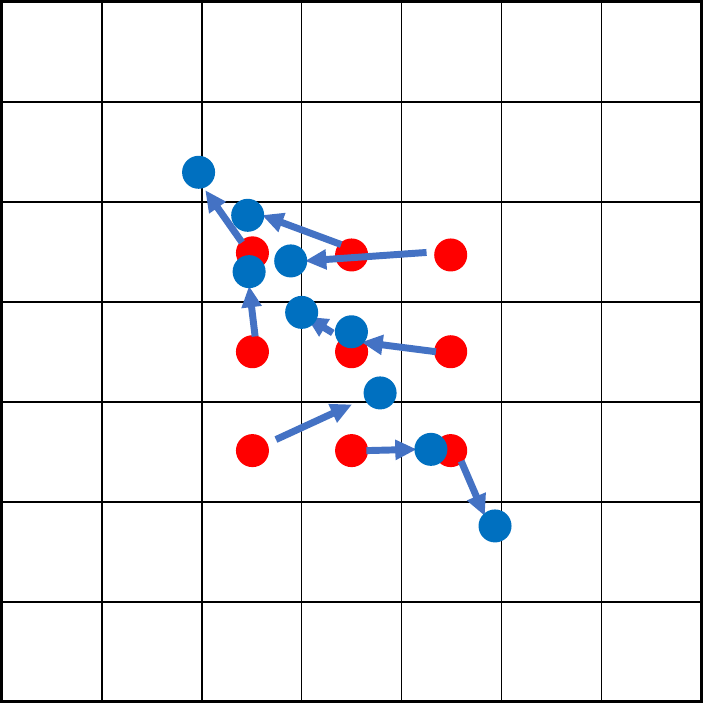} 
  }
\end{center}
\vspace{-1.5ex}
  \caption{Receptive field of different convolution methods: Compared with CNN and DCN, RCN can learn the shape and direction guided offset from the OSH. In the vertex region, the convolution has a rotated cross-star shape sampling points in (c). In the middle of edge area, the reception field shape is inclined straight in (d).} 
  \label{fig7} 
\end{figure}

\vspace{-1ex}
Different from DCN with fully autonomous learning, RCN take the supervised OSH as offset learning guidance. 
The OSH contains both shape and direction information, which could be transformed into sampling coordinate offset through the guided learning. Fig \ref{fig7} depicts the receptive field of different CNN modules. Compared with CNN and DCN, the sampling points of RCN are more consistent with the contour of the object, by which we could extract more valuable features. Experimental results prove that RCN will further improve localization precision.



\vspace{-1.5ex}
\vspace{-2.5ex}

\subsection{Keypoint reorder}
\label{sec3_3}

\begin{figure}[htbp]
\begin{center}
  \subfigure[]{ 
    \label{fig8_1} 
    \includegraphics[height=0.76in]{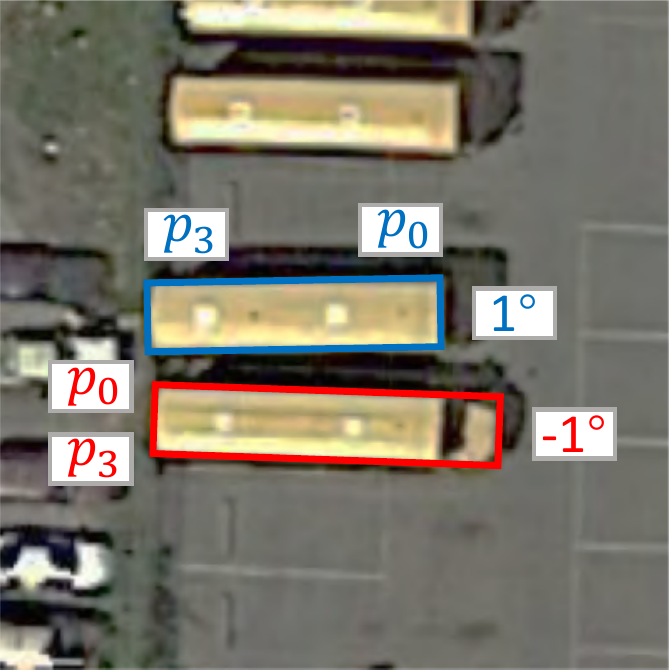}
  }\hspace{-1.08ex}
  \subfigure[]{ 
    \label{fig8_2} 
    \includegraphics[height=0.76in]{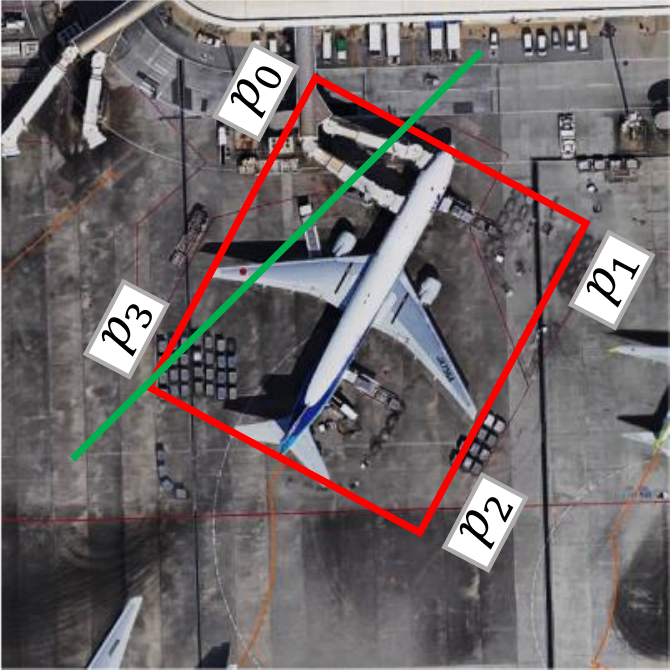}
  }\hspace{-1.08ex}
  \subfigure[]{ 
    \label{fig8_3} 
    \includegraphics[height=0.76in]{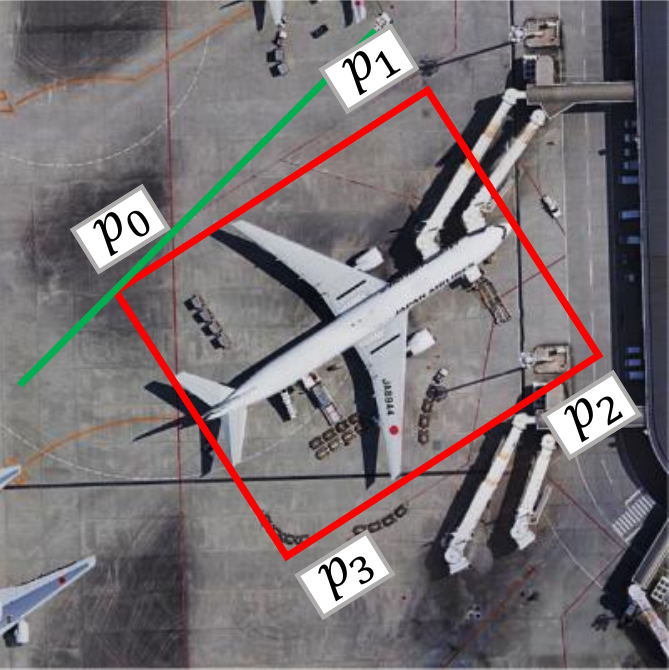}
  }\hspace{-1.1ex}
  \subfigure[]{ 
    \label{fig8_4} 
    \includegraphics[width=0.827in]{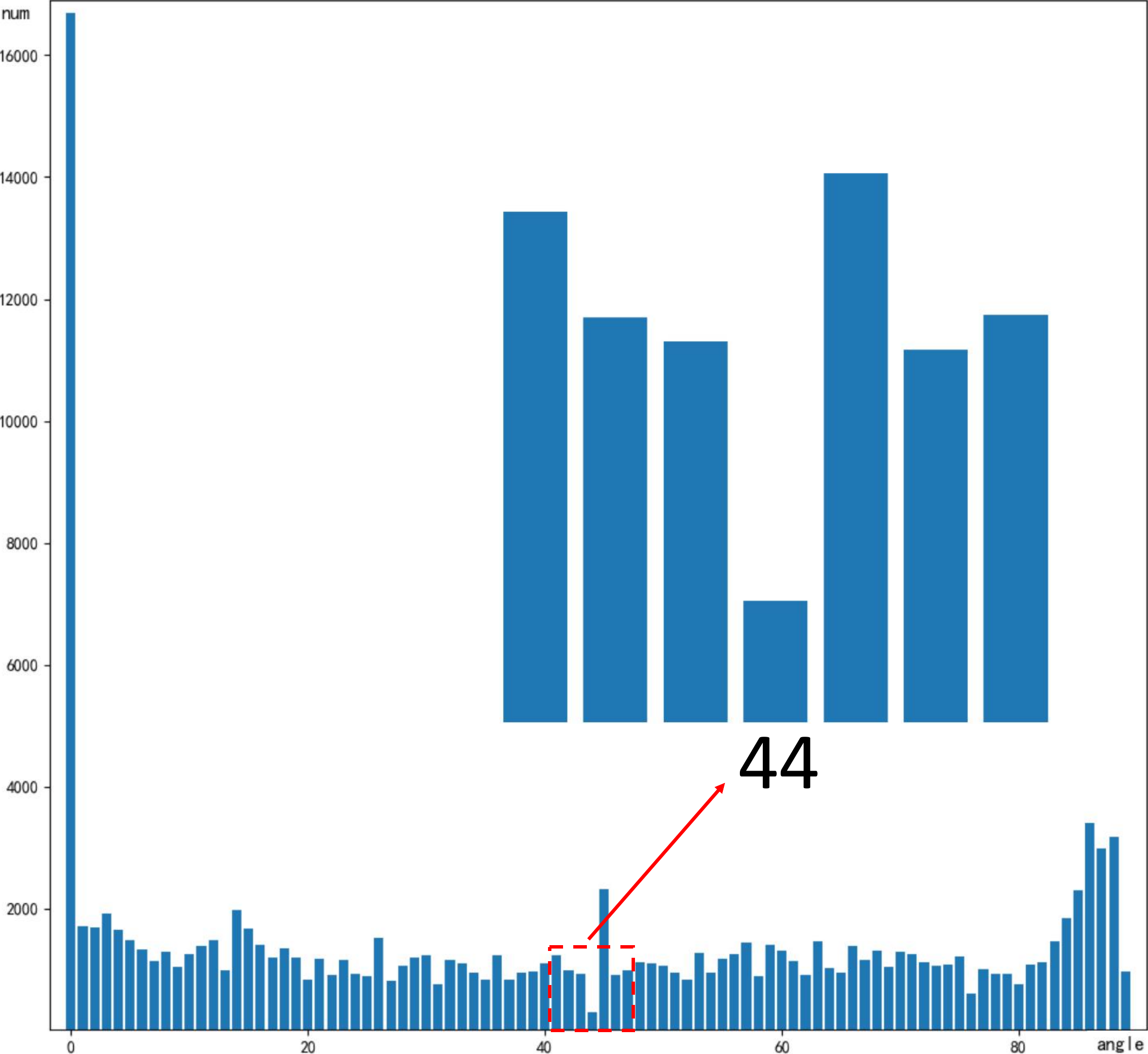}
  }
\end{center} 
\vspace{-1.5ex}
  \caption{(a) Two similar samples with different labels in the 0$^{\circ}$ interface. (b) and (c) show two cases of our method to sort keypoint order, the green line is minimum angle interface, instance across this angle threshold will transform point order. (d) DOTA angle statistics, the instances size near 44$^{\circ}$ is the least.} 
  \label{fig8} 
\end{figure}

The root of keypoint order confusion lies in the point order transform on the keypoint sorting interface. Most works adopt an angle to sort keypoint, e.g. in Gliding Vertex\cite{glide_vertex}, the author defines the vertex with smallest ordinate as the first point, and then clockwise ordering, which means they choose the horizontal 0$^{\circ}$ as the interface. The confusion near 0$^{\circ}$ is largest, as a slight angle spin will cause the point sets order transform, such as the two cars in Fig \ref{fig8_1}. And in any angle, there is such a confusion problem.

Our method is to minimize the confusion, that is, we need to find an angle value that contains as few instances as possible nearby. We study the angular distribution of the training set. We first find the vertex with the smallest ordinate (for horizontal rectangle, we take the top-left point) as the first point $v_{i}$, then reorder it clockwise to find the last point $v_{(i-1)\%4}$, and calculate the angle between $v_{i}$ and $v_{(i-1)\%4}$. Take DOTA\cite{DOTA} dataset as an example, the angle distribution is shown in the Fig \ref{fig8_4}.
\vspace{-1.5ex}
\vspace{-1ex}


For DOTA\cite{DOTA}, the angle statistics in 44-45$^{\circ}$ interval are the smallest, and the predicted error tolerance rate in this interval is relatively highest. The statistics near 0$^{\circ}$ (90$^{\circ}$) is the largest, which means that the confusion here is also the largest. We reorder 4 vertices according to the minimum confusion angle by Algorithm \ref{alg1}, and then 4 midpoints. Fig \ref{fig8_2} and Fig \ref{fig8_3} display two cases of our keypoint sorting algorithm.

\begin{figure}[htbp]
\centering
  \subfigure{ 
    \label{fig9_1} 
    \includegraphics[width=2.72in]{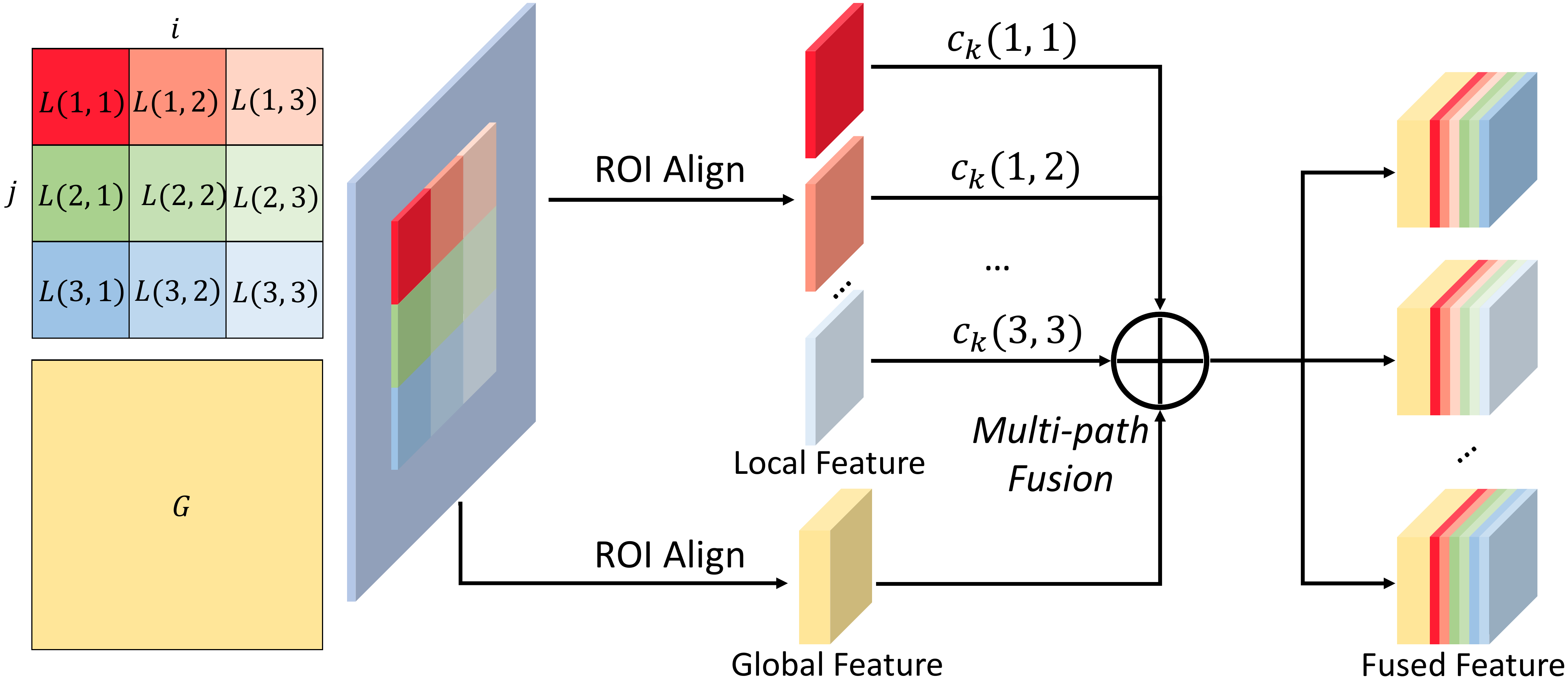}
  }\hspace{-1.5ex}
  \subfigure{ 
    \label{fig9_2} 
    \includegraphics[width=0.48in]{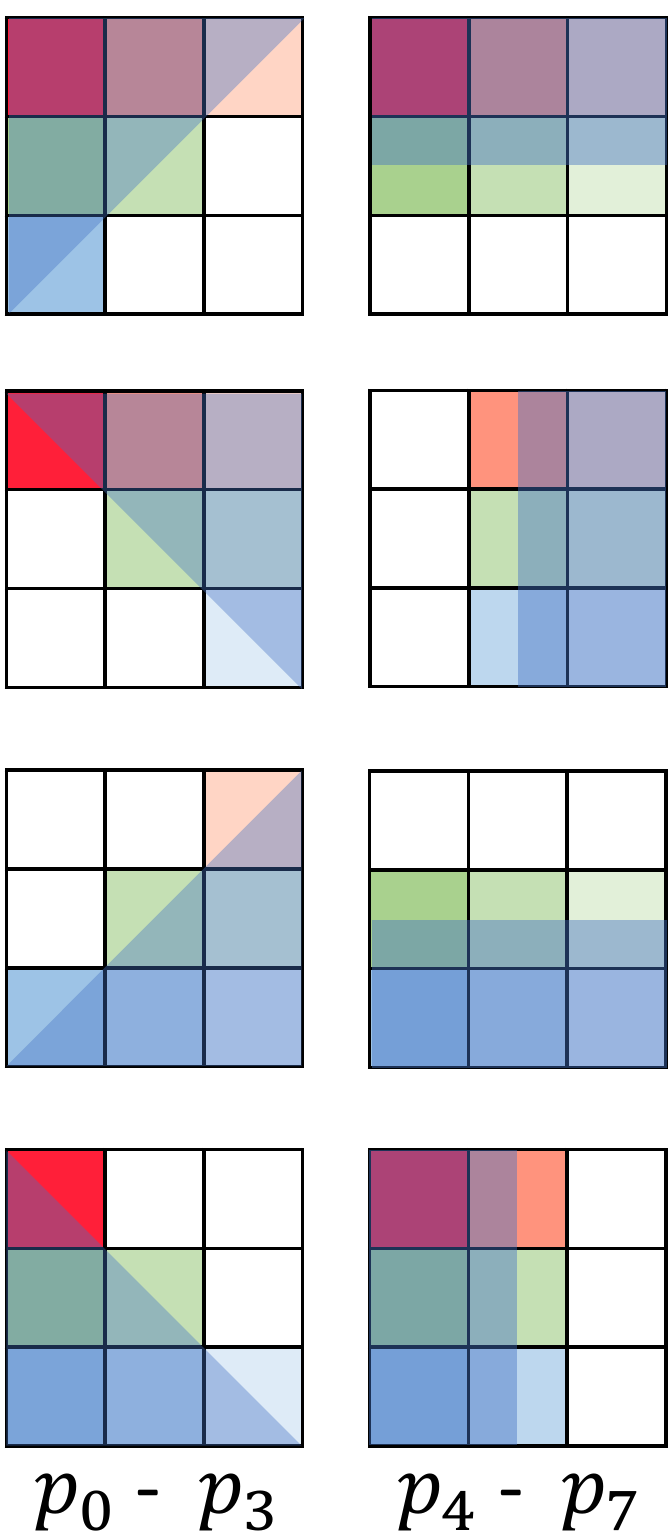} 
  } 
  \vspace{-1.5ex}
\caption{Multi-path feature fusion. Each point has a relatively fixed region, as the colored grids in the box on the right side. We fuse global and local features according to the point location.}
\label{fig9}
\end{figure}

\vspace{-1ex}
\begin{algorithm}[htbp]
\caption{Keypoint Reorder} 
\hspace*{0.02in} {\bf Input:} 
4 vertices $(v_{0}, v_{1}, v_{2}, v_{3})$\vspace{0.5ex}\\
\hspace*{0.02in} {\bf Output:} 
4 ordered keypoints $(p_{0},p_{1},p_{2},p_{3})$
\begin{algorithmic}[1]
\State $V \gets  \{v_{0}, v_{1}, v_{2}, v_{3}\}$;
\State $i = find \ the  \ index \ of \ top \ point(V)$;
\If{$y_{i}=y_{(i-1) \% 4}$} \vspace{0.3ex}
\State $i \gets (i-1) \% 4$;  
\EndIf
\State $angle = argtan\left(\frac{y_{i}-y_{(i-1) \% 4}}{x_{i}-x_{(i-1) \% 4}}\right)$;
\If{$angle > threshold$}
    \State return $v_{i}, v_{(i+1) \% 4}, v_{(i+2) \% 4}, v_{(i+3) \% 4}$;
\Else
    \State return $v_{(i+1) \% 4}, v_{(i+2) \% 4}, v_{(i+3) \% 4}, v_{i}$;
\EndIf
\end{algorithmic}
\label{alg1}
\end{algorithm}
\vspace{-1.5ex}
Each point after reordering has a relatively fixed position, which could provide a more accurate location information. We design a multi-path feature fusion module (MFF). By superimposing global feature and more precise local features on different channels, MMF could utilize the location information more effectively, as shown in Fig \ref{fig9}. In detail, we split the ROI area to $N \times N$ grids ($N$ is set to 3) and define the local feature as $L(i, j)_{i, j \in[1, N]} $. MFF extracts the global and local feature by ROI Align at the same time, and fuse them according to the point rough location. For DOTA, when we choose 44$^{\circ}$ as the keypoint reorder interface, the first keypoint $p_{0}$ can only occur in the left-top triangle region of the ROI, and the local area is (1,1), (1,2), (1,3), (2,1), (2,2) and (3,1), and so are other keypoints, as the color labeled grids in the ROI of Fig \ref{fig9} right side depicts. The $kth$ keypoint feature fusion can be defined by Eq \ref{eq9}. When one grid cell falls into the region where the point is located, the coefficient $c_{k}$ of this grid is 1, otherwise 0.

\vspace{-1ex}

\vspace{-4.2ex}
\begin{equation}\label{eq9}
F_{k}=\sum_{i=1}^{3} \sum_{j=1}^{3} c_{k}(i, j) * L(i, j) + G
\end{equation}
\vspace{-3ex}



\section{Experiments}

\subsection{Dataset and implementation details}

To compare the performance of OSKDet with the state-of-the-art methods, we conduct evaluations on several public datasets, including aerial detection dataset DOTA\cite{DOTA}, HRSC2016\cite{hrsc2016} and UCAS-AOD\cite{ucas-aod}, text detection dataset ICDAR2015\cite{ICDAR2015}  and ICDAR2017MLT\cite{icdar2017}.

\textbf{DOTA}\cite{DOTA} contains 2806 aerial images, ranging in size from 800 * 800 to 4000 * 4000, and contains 188282 instances in 15 categories. Train, validation and test set are split to 3:1:2. Same as other models, we use train set and validation set for training. We adopt the same processing strategy as\cite{roitransformer,glide_vertex}, which cropped the image to 1024*1024 with 824 pixels as a stride. 

\textbf{HRSC2016}\cite{hrsc2016} is a high resolution ship recognition dataset which contains 1061 images, and the image size ranges from 300 * 300 to 1500 * 900.

\textbf{UCAS-AOD}\cite{ucas-aod} contains 1510 aerial images of about 659 * 1280 pixels, with 2 categories (car and plane) of 14,596 instances.

\textbf{ICDAR2015}\cite{ICDAR2015} and \textbf{ICDAR2017MLT}\cite{icdar2017} are two incidental scene Text dataset. Both of them contain natural scene text images with location annotations. IC15 has 1500 images and IC17 has 18000 images.


OSKDet is implemented on pytorch. We use 3 RTX 2080ti GPU, 2 images per GPU in our experiment. We adopt the FPN\cite{fpn} based ResNet50\cite{resnet} and ResNet101 as our backbone. We train 12 epochs in total by using momentum gradient descent optimization. The weight decay is set to 0.0001 and momentum is set to 0.9. The initial learning rate was 0.0075 and divided by 10 at (8, 11) epochs. For aerial set, we resize the image to 1$\times$ and 0.5$\times$ scales and use random rotation (0$^{\circ}$, 90$^{\circ}$, 270$^{\circ}$) in training. For text set, we resize the longer side to \{800, 1000, 1200\}. We also adopt class balancing for DOTA.

\vspace{-0.5ex}
\subsection{Aerial detection}

\textbf{DOTA.} Tab \ref{table_dota} shows our testing results on the DOTA v1.0 OBB task. Compared with the state-of-the-art methods, OSKDet shows superior performance. With ResNet50 as backbone, OSKDet achieves 76.22\% AP, which excels most models with deeper feature extractor. With ResNet101 as backbone, OSKDet achieves 77.81\% AP, which outperforms all previous works. Compared with the angle based detectors\cite{rrpn,csl,scrdet++}, OSKDet improves the state-of-the-art method by 1.00\% AP. Compared with other vertex based detectors\cite{polardet,glide_vertex,rsdet}, OSKDet improves the state-of-the-art method by 1.17\% AP. Compared with the vertex offset regression method\cite{glide_vertex}, OSKDet improves AP by 2.79\%. Furthermore, OSKDet has achieved the best performance in multiple categories, especially in irregularly shaped categories such as baseball diamond, ship and helicopter, which illustrates the great advantages of OSKDet in spatial representation and transformation. Through accurate characterization and feature extraction of keypoints, OSKDet is more effective in detecting complex and irregular shape targets. 

\textbf{HRSC2016 and UCAS-AOD.} Tab \ref{hrsc_1} depicts the comparison of different methods on HRSC2016\cite{hrsc2016} and UCAS-AOD\cite{ucas-aod}. OSKDet achieves 89.91\% AP and 97.18\% AP respectively in the two challenging datasets and outperforms all other methods, which demonstrates the proposed OSKDet brings consistent gain on different aerial datasets. 
\vspace{-0.5ex}
\begin{table}[htbp]

\begin{minipage}[b]{.43\linewidth}
\centering
\begin{scriptsize}
\begin{tabular}{l|l}
\toprule
Method                              & AP \\ 
\hline
RoI-Trans\cite{roitransformer}      &86.20      \\
RSDet\cite{rsdet}                   &86.50      \\
Gliding Vertex\cite{glide_vertex}     &88.20      \\
BBAVectors\cite{bba}                &88.60       \\
CSL\cite{csl}                       &89.62      \\ 
OSKDet(ours)                         &\textbf{89.91}   \\\bottomrule   
\end{tabular}
\end{scriptsize}

\end{minipage}\begin{minipage}[b]{.57\linewidth}

\centering
\begin{scriptsize}
\begin{tabular}{l|p{12pt}p{12pt}p{12pt}}
\toprule
Method                       & Plane & Car   & AP   \\ \hline

S2ARN\cite{bao2019single}    & 97.60 & 92.20 & 94.90 \\
FADet\cite{fadet}            & 98.69 & 92.72 & 95.71 \\
R3Det\cite{r3det}            & 98.20 & 94.14 & 96.17 \\
SCRDet++\cite{scrdet++}      & 98.93 & 94.97 & 96.95 \\
PolarDet\cite{polardet}      & \textbf{99.08} & 94.96 & 97.02 \\ 
OSKDet(ours)                  & 99.07 & \textbf{95.29} & \textbf{97.18}   \\ \bottomrule
\end{tabular}
\end{scriptsize}

\end{minipage}
\vspace{0.5ex}
\caption{Comparison with state-of-the-art methods on HRSC2016 (left) and UCAS-AOD (right).}
\label{hrsc_1}
\end{table}

\vspace{-1.2ex}

\textbf{Ablation study.} We perform all ablation experiments on DOTA dataset. Unless otherwise specified, all test benchmarks are VOC2007\cite{voc} AP.5 IOU metric. Tab \ref{table11} shows the comprehensive ablation experiment results.
\vspace{-1ex}
\begin{table}[htbp]
  \centering
  \begin{scriptsize}
  \begin{tabular}{l|lllll|l|l}
  \toprule
  Method     & MS            &OSH             &RCN           &KR          &MFF   & AP.5   &FPS \\ \hline
  \multirow{9}{*}{OSKDet}          &        &          &        &       &                  & 73.38      & -       \\
             & \checkmark    &                &              &             &               & 75.54     &  -    \\
             & \checkmark    &\checkmark       &             &             &               & 76.16      & 8.15       \\
             & \checkmark    &\checkmark       & \checkmark  &             &               &  76.86   &  7.91   \\
             & \checkmark    &                &             & \checkmark   &               &  76.10  &   -    \\
             & \checkmark    &                &             & \checkmark    & \checkmark   &  76.75   &  6.88     \\
             & \checkmark    &\checkmark       &             & \checkmark   &              &  76.66   &  -     \\
             &             &\checkmark       & \checkmark    & \checkmark   & \checkmark   &  76.06    & 6.13    \\
             & \checkmark  &\checkmark & \checkmark &\checkmark     & \checkmark    & \textbf{77.81}   & -     \\ \bottomrule
  \end{tabular}
  \end{scriptsize}
  \vspace{0.7ex}
  \caption{Ablation experiment of different strategies on DOTA. MS, OSH, RCN, KR and MFF mean multi-scale training and testing, orientation-sensitive heatmap, rotation-aware deformable convolution, keypoint reorder and multi-path feature fusion respectively.}
  \label{table11}
  \end{table}

\vspace{-1ex}

\begin{table*}[]


\centering
\begin{scriptsize}
\begin{tabular}{l|l|p{12pt}p{12pt}p{12pt}p{12pt}p{12pt}p{12pt}p{12pt}p{12pt}p{12pt}p{12pt}p{12pt}p{12pt}p{12pt}p{12pt}p{12pt}p{12pt}}
\toprule
Method            & Backbone          & PL    & BD    & BR    & GTF   & SV    & LV    & SH    & TC    & BC    & ST    & SBF   & RA    & HA    & SP    & HC    & AP.5  \\ 
\hline
FR-O\cite{DOTA}       & ResNet101 & 79.09 & 69.12 & 17.17 & 63.49 & 34.20  & 37.16 & 36.20  & 89.19 & 69.6  & 58.96 & 49.40  & 52.52 & 46.69 & 44.80  & 46.30  & 52.93 \\
IENet\cite{ienet}      & ResNet101         & 80.20  & 64.54 & 39.82 & 32.07 & 49.71 & 65.01 & 52.58 & 81.45 & 44.66 & 78.51 & 46.54 & 56.73 & 64.40  & 64.24 & 36.75 & 57.14 \\
PIOU\cite{piou}       & DLA-34    & 80.90  & 69.70  & 24.10  & 60.20  & 38.30  & 64.40  & 64.80  & 90.90  & 77.20  & 70.40  & 46.50  & 37.10 & 57.10  & 61.90  & 64.00    & 60.50  \\
R2CNN\cite{r2cnn}       & ResNet101         & 80.94 & 65.67 & 35.34 & 67.44 & 59.92 & 50.91 & 55.81 & 90.67 & 66.92 & 72.39 & 55.06 & 52.23 & 55.14 & 53.35 & 48.22 & 60.67 \\
RRPN\cite{rrpn}        & ResNet101         & 88.52 & 71.20  & 31.66 & 59.30  & 51.85 & 56.19 & 57.25 & 90.81 & 72.84 & 67.38 & 56.69 & 52.84 & 53.08 & 51.94 & 53.58 & 61.01 \\
ROI-Trans\cite{roitransformer}  & ResNet101         & 88.64 & 78.52 & 43.44 & 75.92 & 68.81 & 73.68 & 83.59 & 90.74 & 77.27 & 81.46 & 58.39 & 53.54 & 62.83 & 58.93 & 47.67 & 69.56 \\
RSDet\cite{rsdet}      & ResNet152         & 90.10  & 82.00    & 53.80  & 68.50  & 70.20  & 78.70  & 73.60  & \textbf{91.20}  & 87.10  & 84.70  & 64.30  & 68.20  & 66.10  & 69.30  & 63.70  & 74.10  \\
Gliding Vertex\cite{glide_vertex} & ResNet101          & 89.64 & 85.00    & 52.26 & 77.34 & 73.01 & 73.14 & 86.82 & 90.74 & 79.02 & 86.81 & 59.55 & 70.91 & 72.94 & 70.86 & 57.32 & 75.02 \\
FFA\cite{ffa}        & ResNet101         & 90.10  & 82.70  & 54.20  & 75.20  & 71.00   & 79.90  & 83.50  & 90.70  & 83.90  & 84.60  & 61.20  & 68.00    & 70.70  & 76.00    & 63.70  & 75.70  \\
CenterMap\cite{centermap} & ResNet101         & 89.83 & 84.41 & 54.60  & 70.25 & 77.66 & 78.32 & 87.19 & 90.66 & 84.89 & 85.27 & 56.46 & 69.23 & 74.13 & 71.56 & 66.06 & 76.03 \\
CSL\cite{csl}        & ResNet152         & \textbf{90.25} & 85.53 & 54.64 & 75.31 & 70.44 & 73.51 & 77.62 & 90.84 & 86.15 & 86.69 & \textbf{69.66} & 68.04 & 73.83 & 71.10  & 68.93 & 76.17 \\
PolarDet\cite{polardet}           & ResNet101         & 89.65 & 87.07 & 48.14 & 70.97 & \textbf{78.53} & \textbf{80.34} & 87.45 & 90.76 & 85.63 & 86.87 & 61.64 & 70.32 & 71.92 & \textbf{73.09} & 67.15 & 76.64 \\
SCRDet++\cite{scrdet++}  & ResNet101         & 90.05 & 84.39 & \textbf{55.44} & 73.99 & 77.54 & 71.11 & 86.05 & 90.67 & \textbf{87.32} & \textbf{87.08} & 69.62 & 68.90  & 73.74 & 71.29 & 65.08 & 76.81 \\ \hline
OSKDet(Ours)               & ResNet50  & 89.83 & 84.75 & 52.18 & 75.22 & 71.75 & 76.47 & 87.29 & 90.42 & 78.95 & 87.02 & 57.49 & 69.71 & 73.50  & 71.23 & 60.14 & 75.06 \\
OSKDet-MS(Ours)               & ResNet50  & 89.98 & 86.99 & 53.13 & 75.55 & 72.87 & 76.97 & 87.63 & 90.74 & 78.87 & 86.97 & 60.13 & 70.68 & \textbf{75.70}  & 71.53 & 65.60 & 76.22 \\
OSKDet(Ours)              & ResNet101         & 89.88 & 86.26 & 53.42 & 74.40  & 72.23 & 76.80 & 87.34 & 90.81 & 78.98 & 86.39 & 57.97 & 71.59 & 75.60 & 72.20 & 67.07 & 76.06 \\
OSKDet-MS(Ours)               & ResNet101         & 90.04 & \textbf{87.25} & 54.41 & \textbf{79.48} & 72.66 & 80.29 & \textbf{88.20} & 90.84 & 83.91 & 86.90  & 63.39 & \textbf{71.76} & 75.63 & 72.59 & \textbf{69.75} & \textbf{77.81} \\
\bottomrule

\end{tabular}
\end{scriptsize}
\centering
\vspace{0.1ex}
\caption{Comparison of different method results on DOTA-v1.0 OBB task (MS means multi-scale training and testing)}
\label{table_dota}
\end{table*}

\vspace{-0.5ex}
\textbf{Orientation-sensitive heatmap.} We compare the proposed method with different predicting formats, including vertex offset regression\cite{glide_vertex} (REG), single point heatmap\cite{maskrcnn} (SPH), cross-star heatmap\cite{gridrcnn} (CSH), standard gaussian heatmap\cite{cornernet} (SGH) and orientation-sensitive heatmap (OSH).
We implement the four different format heatmaps on OSKDet. Tab \ref{table5} displays the result. The OSH surpasses all baselines with an increment of 0.62\%-1.15\%. Compared with the SGH, the OSH improves 0.68\% and 0.62\% AP without/ with FPN respectively, which proves the stronger spatial modeling capabilities of OSH will bring localization accuracy gains. 

\begin{table}[htbp]
\centering
\begin{scriptsize}
\begin{tabular}{l|l|l}
\toprule
Method                          & FPN          & AP.5 \\ \hline
REG\cite{glide_vertex}  &              & 73.39        \\
SGH\cite{cornernet}   &        & 73.86(+0.47)     \\
OSH(Ours)     &              & 74.54(+1.15)     \\
REG\cite{glide_vertex}  &  \checkmark  & 75.02        \\
SPH\cite{maskrcnn}            &  \checkmark  & 75.11        \\
CSH\cite{gridrcnn}               &  \checkmark  & 75.42        \\
SGH\cite{cornernet}       &  \checkmark  & 75.54(+0.52)     \\
OSH(Ours)             &  \checkmark  & \textbf{76.16(+1.14)}     \\ \bottomrule
\end{tabular}
\end{scriptsize}
\vspace{0.5ex}
\caption{Comparison of different prediction format results}
\label{table5}
\end{table}

\vspace{-1ex}

\begin{figure}[htbp]
    \centering
    \includegraphics[width=0.8\linewidth]{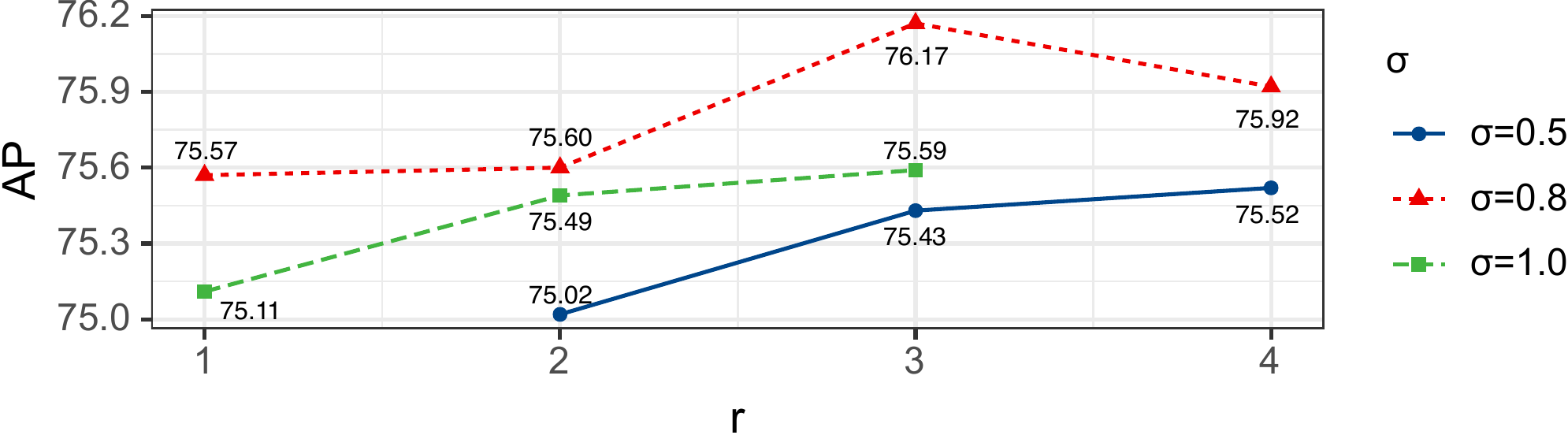}
    \caption{Comparison of different $\sigma$ and $r$ results on OSH. The two parameters represent the direction of rotated object. More accurate characterization will improve the localization accuracy, which demonstrates the superiority of OSH.} 
    \label{fig13}    
\end{figure}

To further verify the effect of OSH and find best parameters, we try different gaussian kernel ratio and kernel size combination experiment. As shown in Fig \ref{fig13}, $\sigma$ represents the standard deviation in the vertical direction $\sigma_{22}$, and $r$ represents the scale factor between the axis direction and the vertical direction, that is, $\sigma_{11}$ = $r * \sigma_{22}$ ($r = 1$ means SGH). We get the best effect when $r = 3$ and $\sigma = 0.8$.


\vspace{-1ex}

\begin{figure}[htbp]
    \centering
    \includegraphics[width=0.85\linewidth]{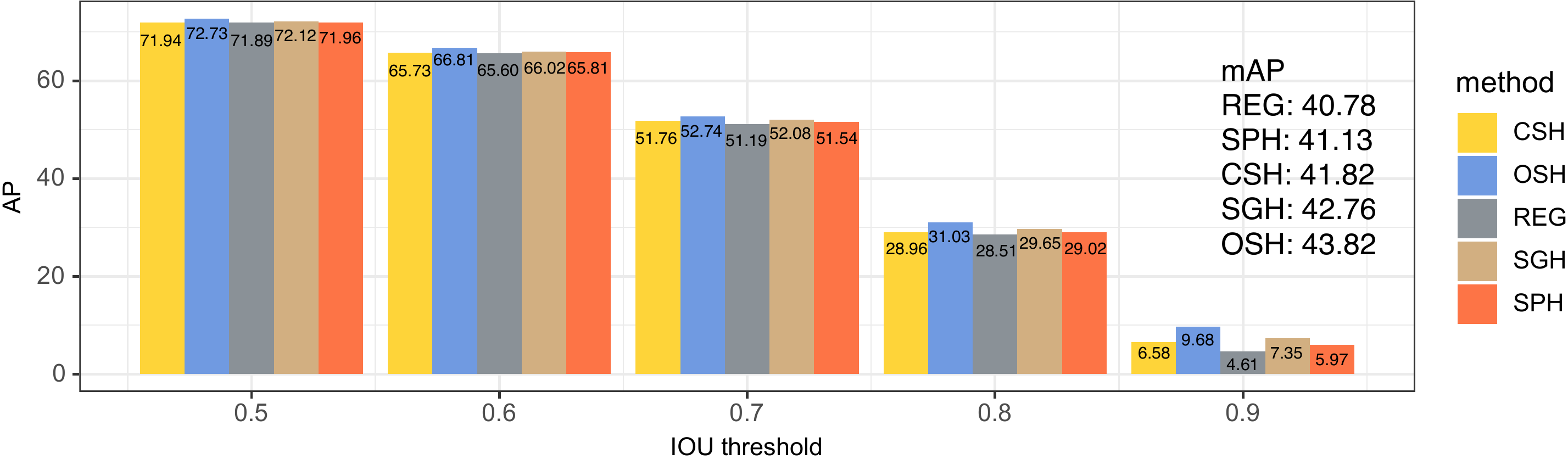}
    \caption{Comparison of different prediction format results under AP.5-.95 IOU metric on validation set}  
    \label{fig14}   
\end{figure}

Since the DOTA\cite{DOTA} official evaluation system only provides the test results for AP.5 metric. In order to test the performance of the model under different IOU metrics, we use the training set to train and validation set to evaluate. We reimplement \cite{glide_vertex} and other format heatmap in OSKDet. The OSH achieves 43.82\% mAP, which excels other format predictions with considerable performance gains, specifically 3.04\%, 2.69\%, 2.00\% and 1.06\% mAP for REG, SPH, CSH and SGH.  As illustrated in Fig \ref{fig14}, OSH surpasses all other method especially in the high IOU metric. Under AP.9 IOU threshold, OSH surpasses other methods by 2.33\%-5.07\%, which demonstrates that OSKDet has a huge advantage in obtaining high quality detections.

\vspace{-1ex}
\vspace{-0.2ex}

\textbf{Rotation-aware deformable convolution.} 
We compare the results of standard CNN, DCN, and RCN under AP.5-.95 metric. Same as the previous section, we use the training set to train and the validation set to test. The RCN improves the mAP by 1.27\% and 0.75\% respectively as shown in Tab \ref{table8}. Through the guided learning of the OSH, RCN can extract highly effective features and obtain better localization results. As shown in Fig \ref{fig11}, when two targets are very close, standard CNN and DCN cannot distinguish them, while the more precise sampling points of RCN can accurately locate them respectively.
\vspace{-1.5ex}
\begin{table}[htbp]
  \centering
  \begin{scriptsize}
  \begin{tabular}{l|ll|lll}
  \toprule
  Method        &   DCN   & RCN  & AP.5  & AP.75 & AP.5-.95 \\ \hline
  
                &      &           & 72.09 & 42.98 & 42.86       \\
  OSKDet          & \checkmark      &   & 72.35 & 43.70 & 43.38(+0.52)     \\
                  &      &\checkmark    & 72.84 & 44.52 & \textbf{44.13(+1.27)}     \\ \bottomrule
  \end{tabular}
  \end{scriptsize}
  \vspace{0.8ex}
  \caption{Comparison of different CNN module results under AP.5-.95 IOU metric on validation set}
  \label{table8}
\end{table}
\vspace{-1ex}

\begin{figure*}[htbp]
  \begin{center}
    \subfigure[]{ 
      \begin{minipage}[t]{0.15\linewidth}
      \label{fig10_1} 
      \includegraphics[height=1.1in]{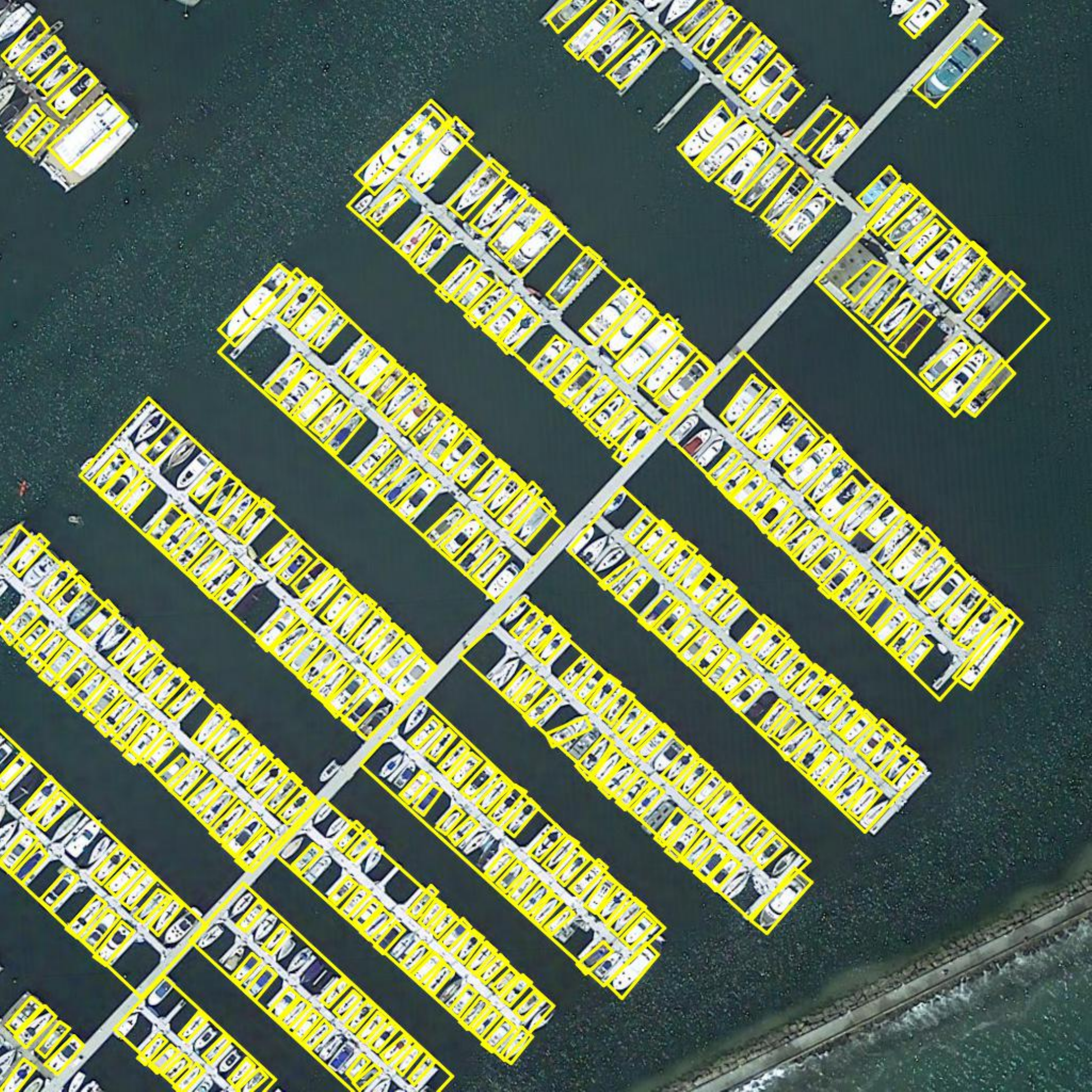} 
      \vspace{2pt}
      \includegraphics[height=1.1in]{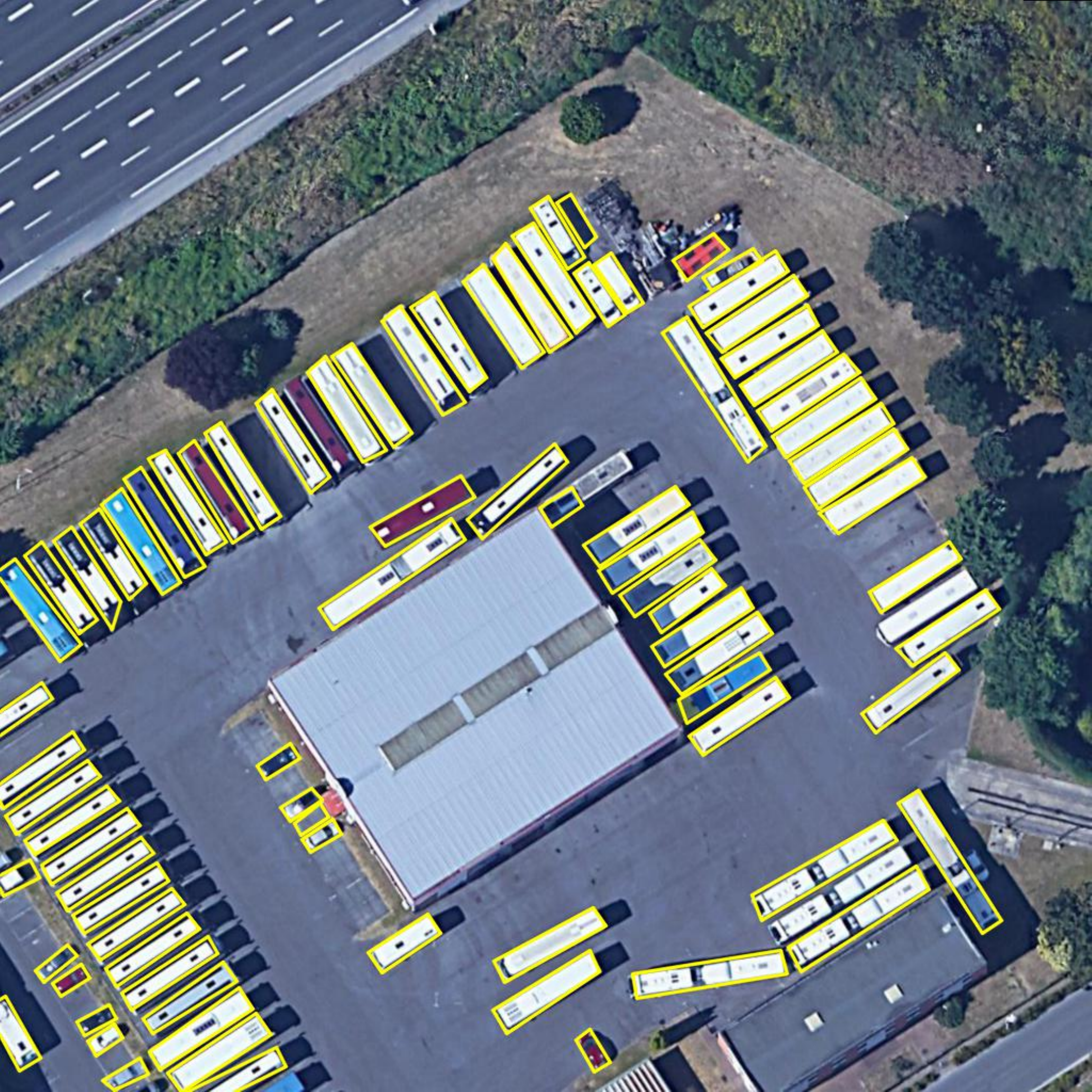} 
      \end{minipage}
    }
    \subfigure[]{ 
      \begin{minipage}[t]{0.15\linewidth}
      \label{fig10_2} 
      \includegraphics[height=1.1in]{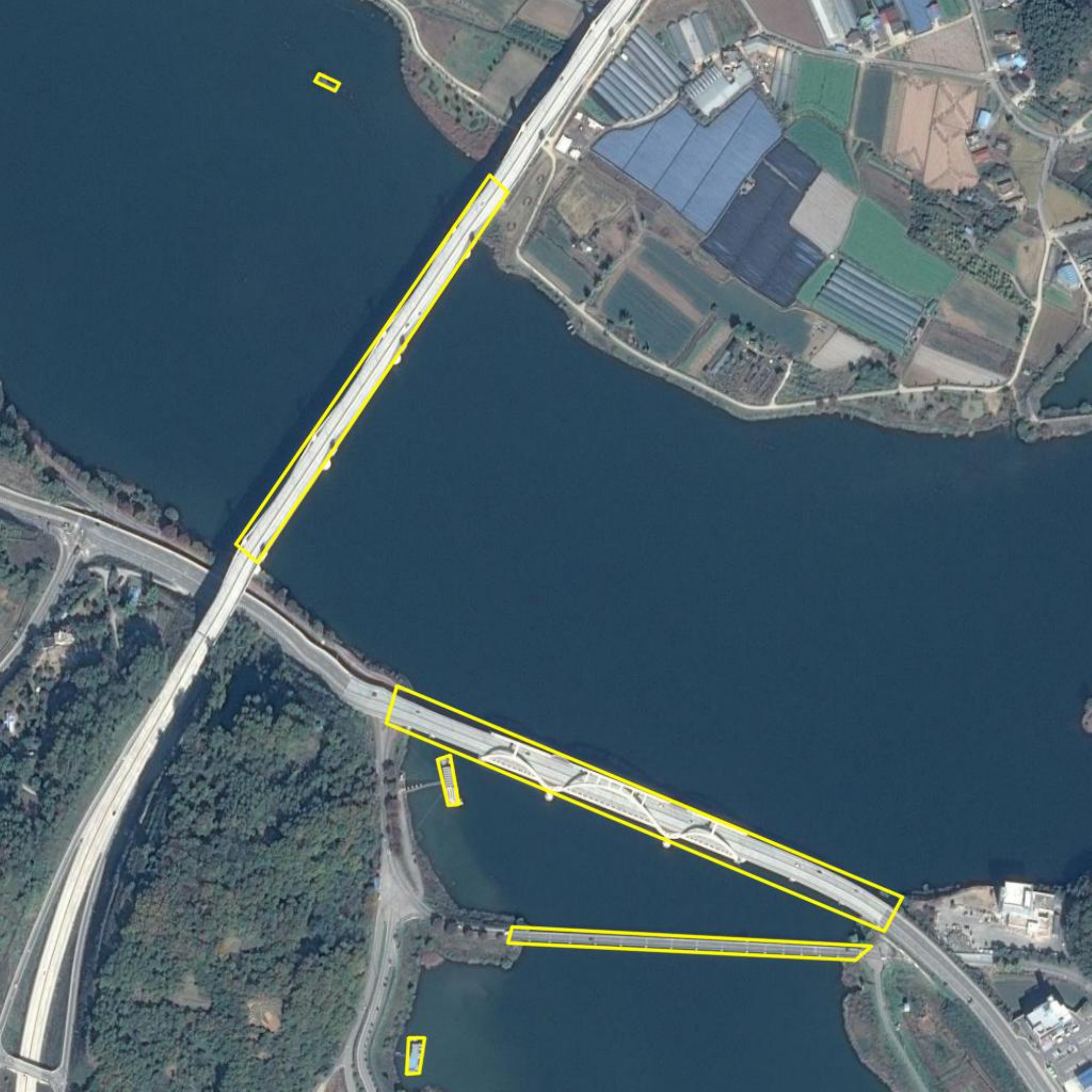} 
      \vspace{2pt}
      \includegraphics[height=1.1in]{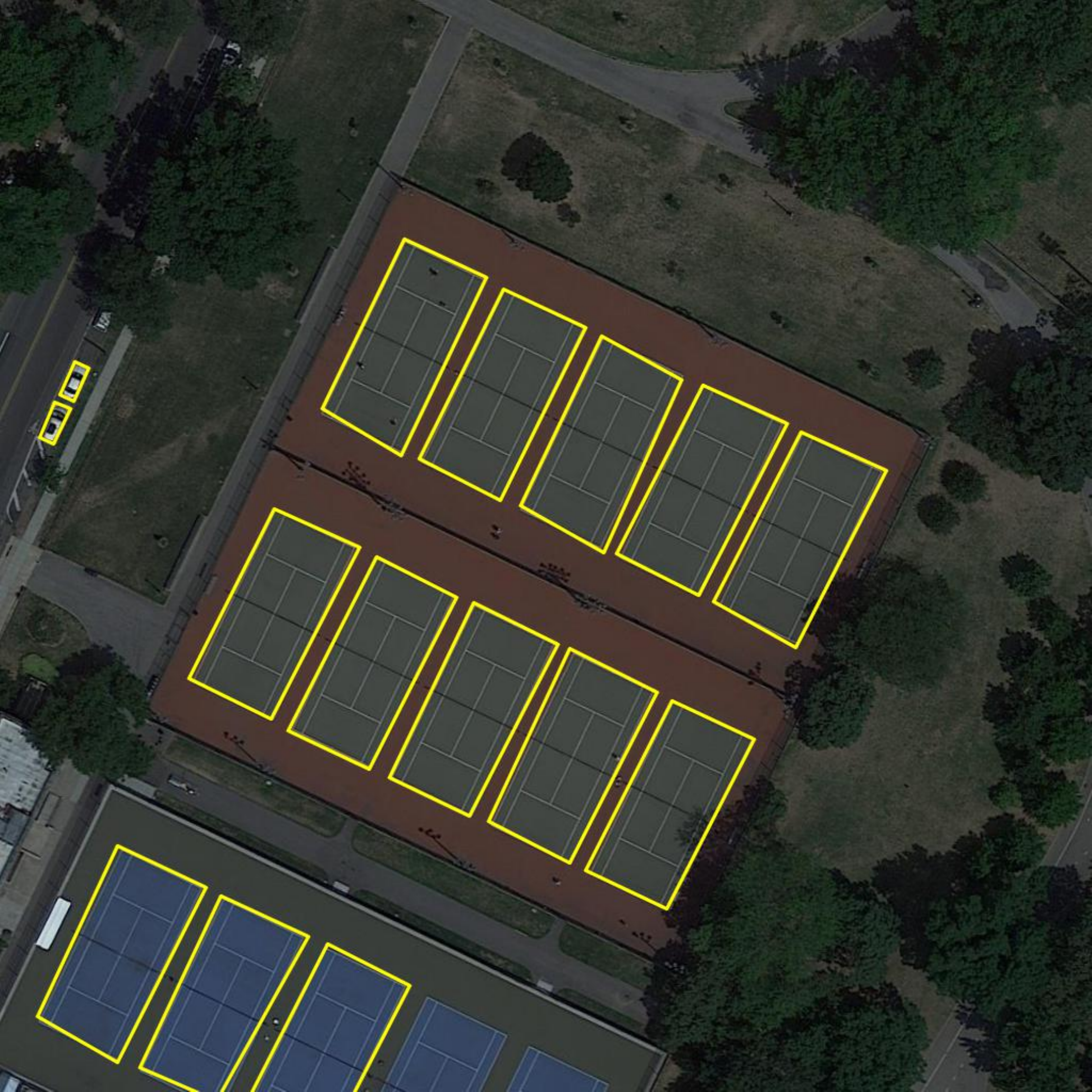} 
      \end{minipage}
    }
    \subfigure[]{ 
      \begin{minipage}[t]{0.15\linewidth}
      \label{fig10_3} 
      \includegraphics[height=1.1in]{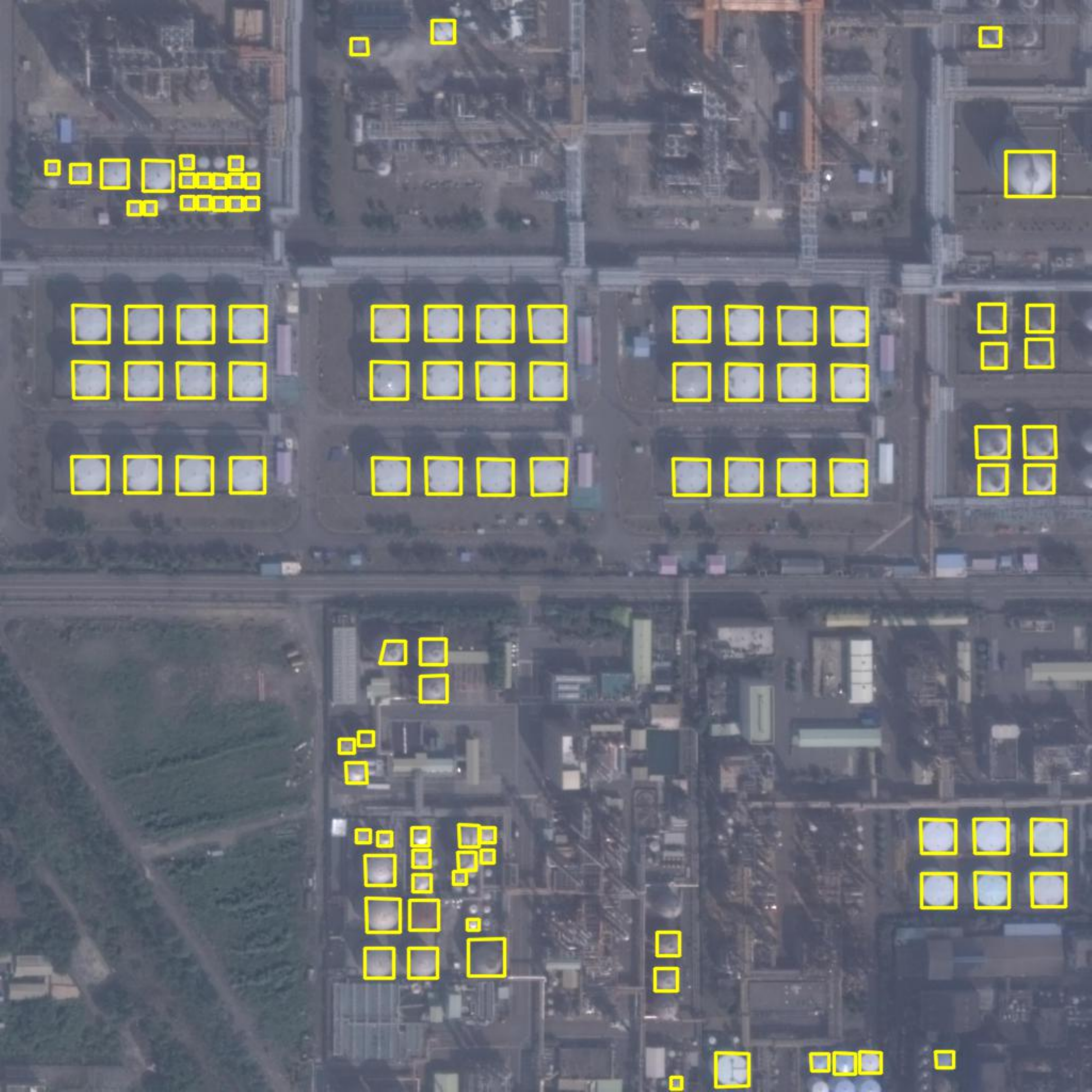} 
      \vspace{2pt}
      \includegraphics[height=1.1in]{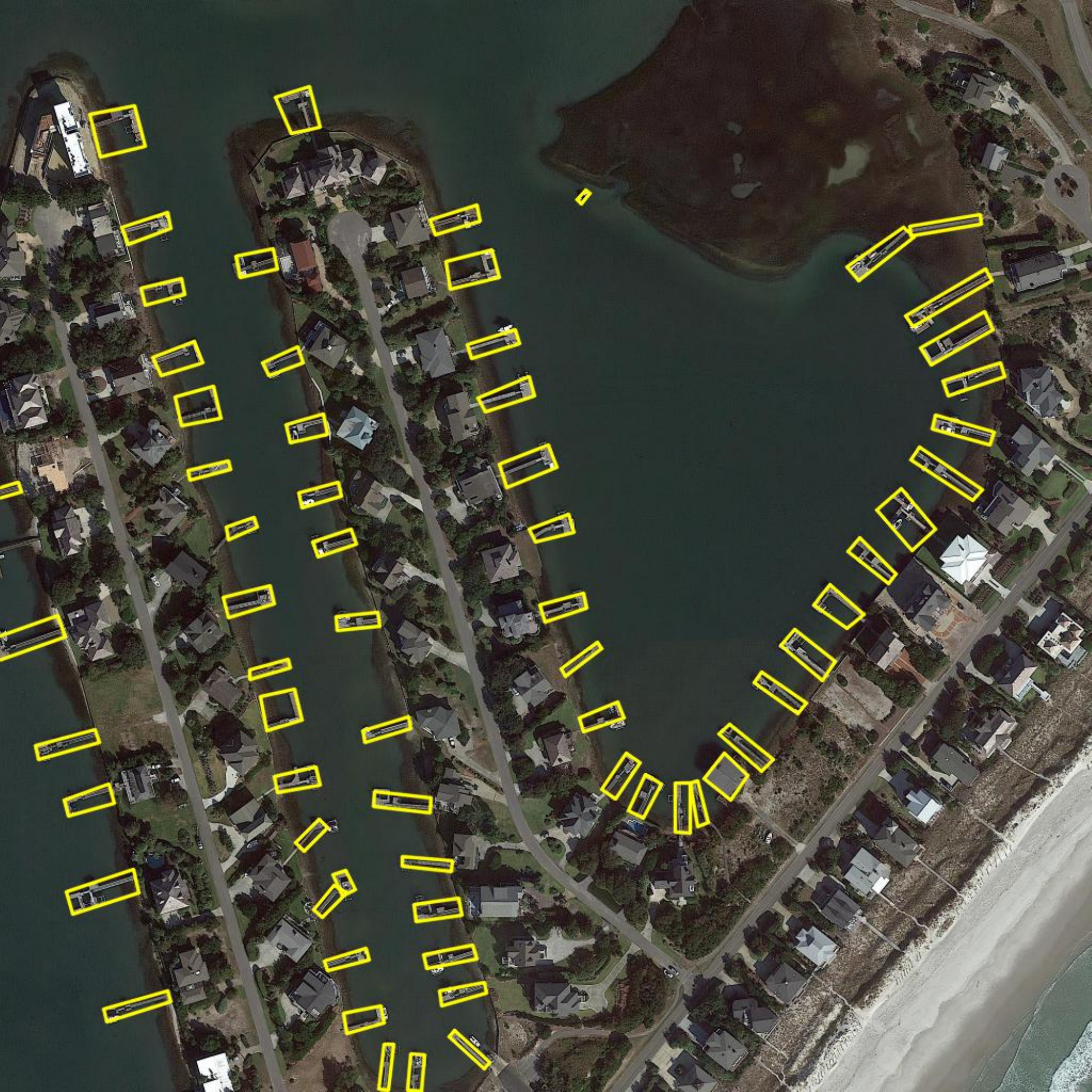} 
      \end{minipage}
    }
    \subfigure[]{ 
      \begin{minipage}[t]{0.15\linewidth}
      \label{fig10_5} 
      \includegraphics[height=1.1in]{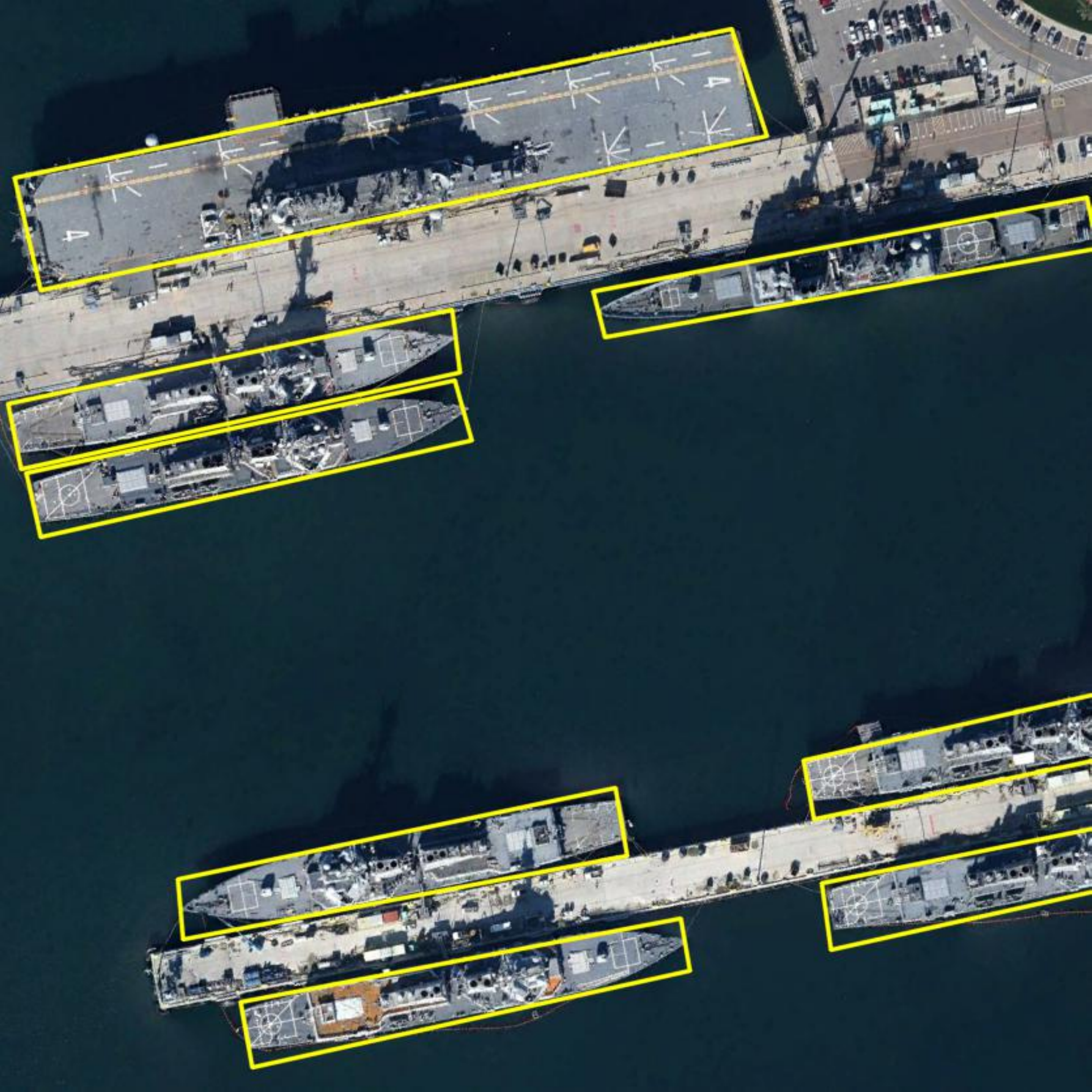} 
      \vspace{2pt}
      \includegraphics[height=1.1in]{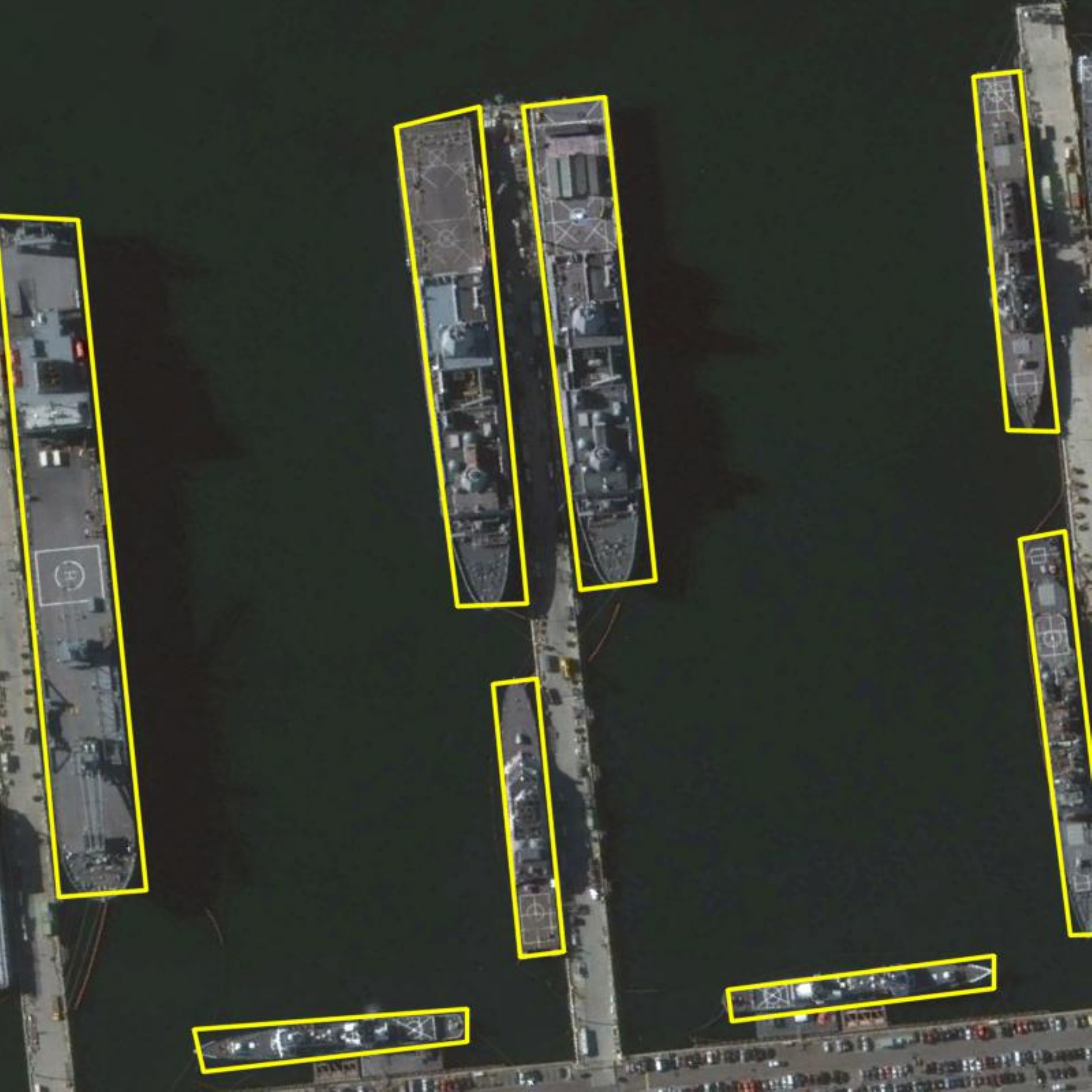} 
      \end{minipage}
    }
    \subfigure[]{ 
      \begin{minipage}[t]{0.15\linewidth}
      \label{fig10_4} 
      \includegraphics[height=1.1in]{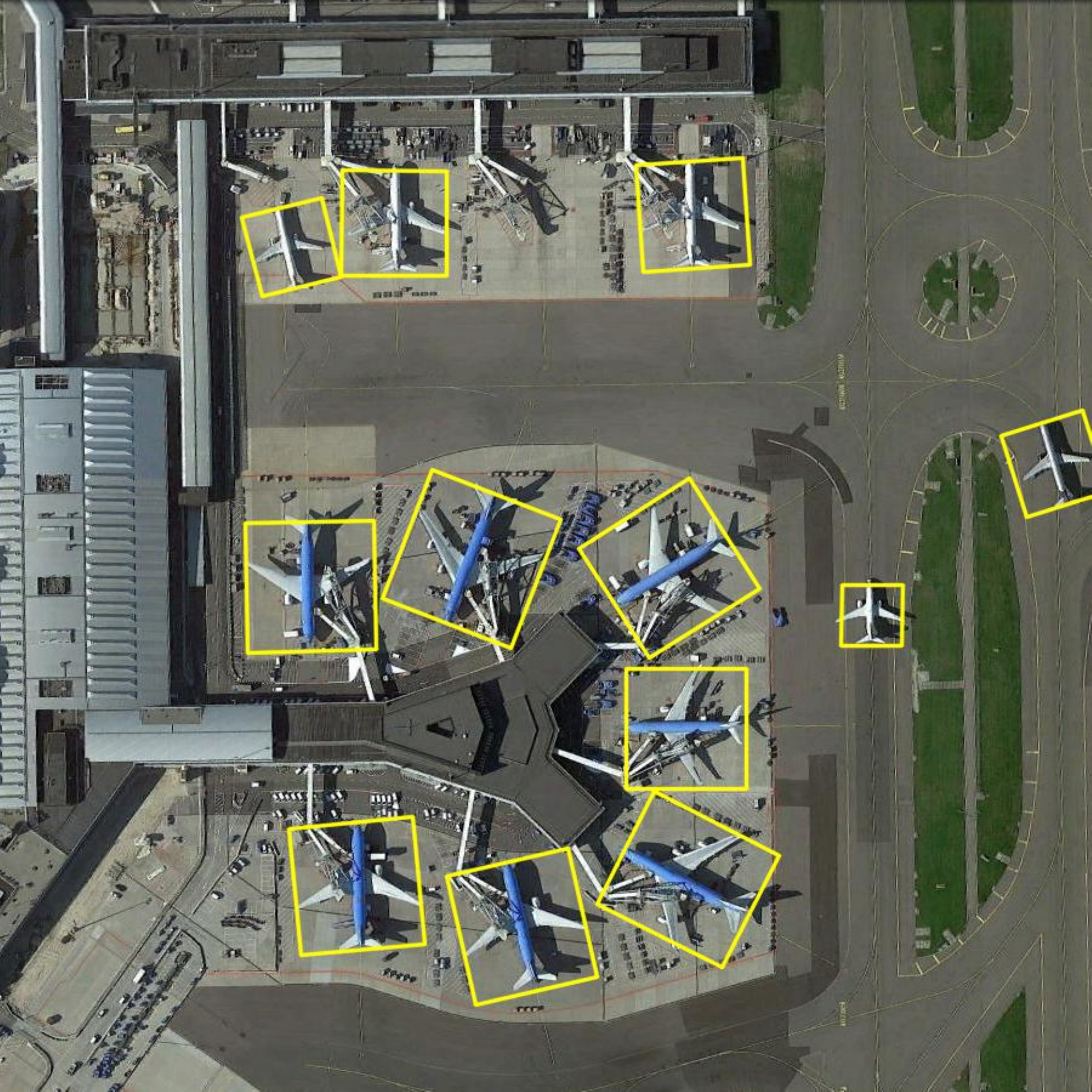} 
      \vspace{2pt}
      \includegraphics[height=1.1in]{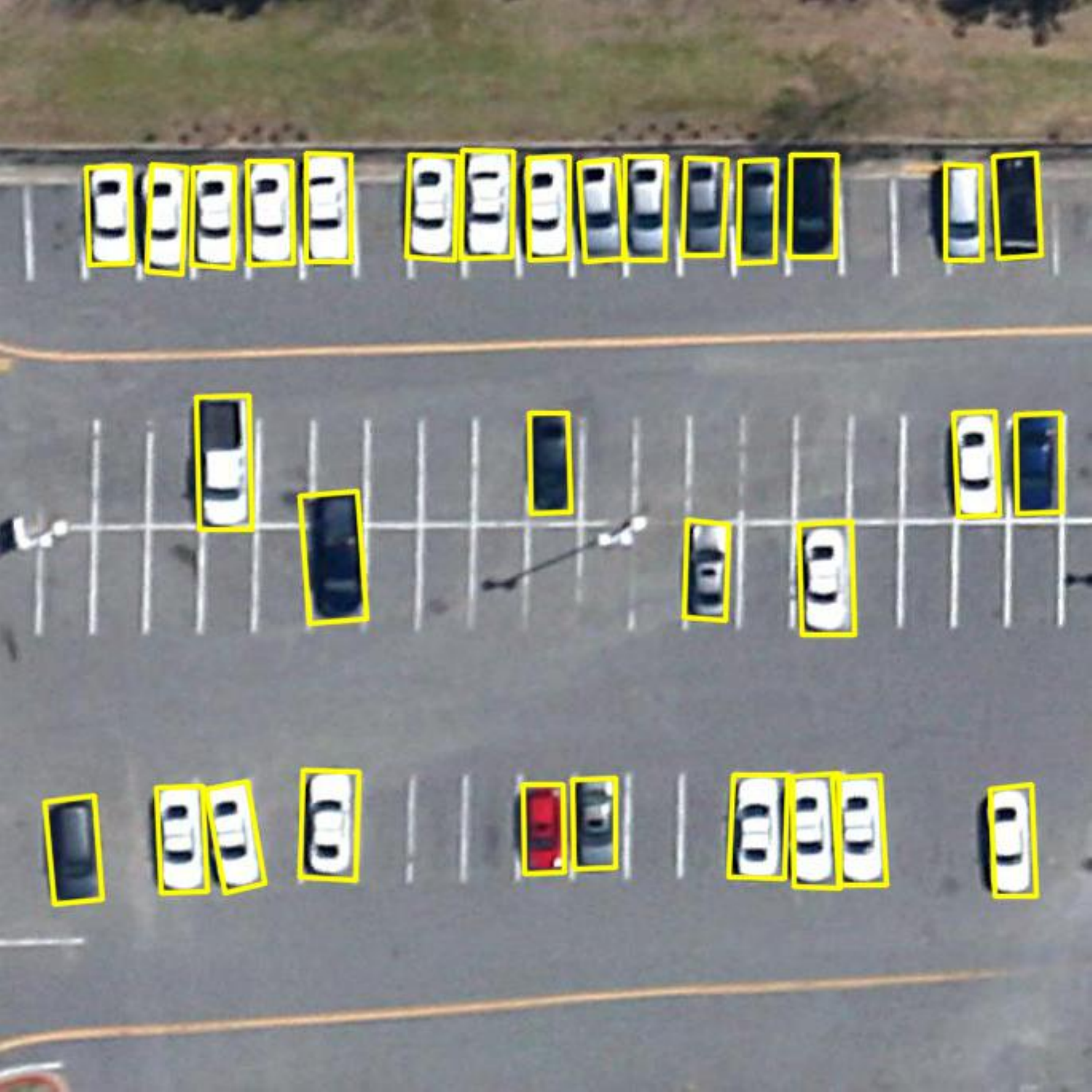} 
      \end{minipage}
    }
    \subfigure[]{ 
      \begin{minipage}[t]{0.15\linewidth}
      \label{fig10_6} 
      \includegraphics[height=1.1in]{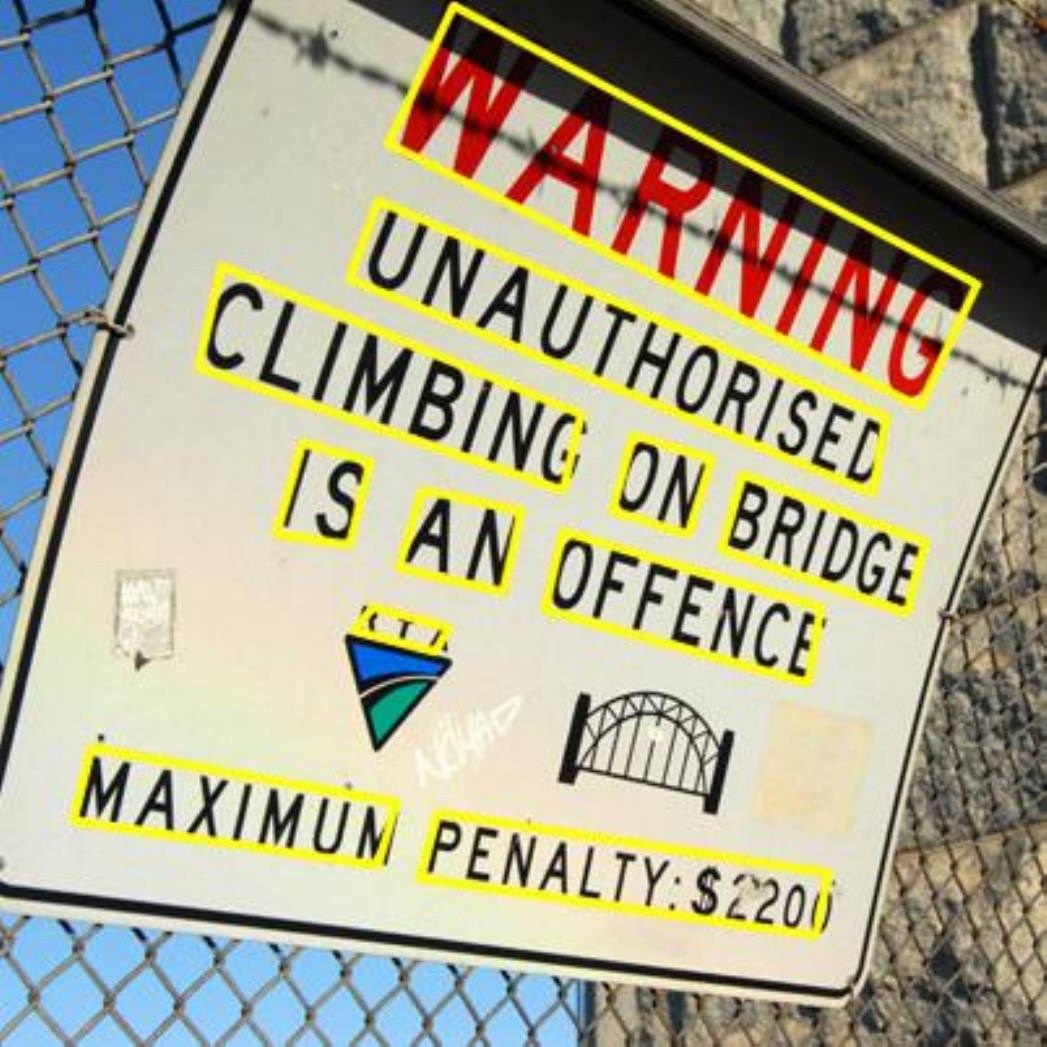} 
      \vspace{2pt}
      \includegraphics[height=1.1in]{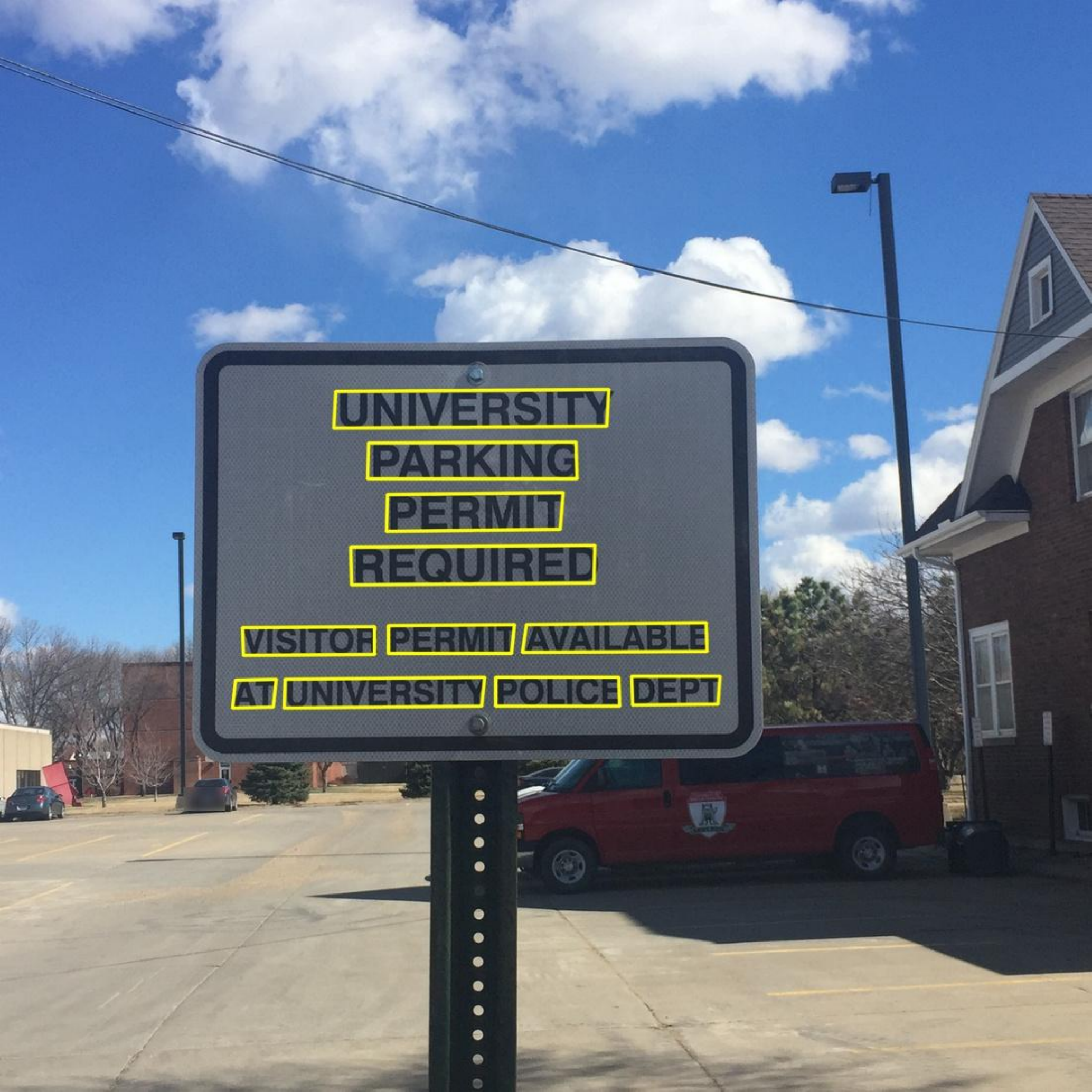} 
      \end{minipage}
    }
  \end{center}
  \vspace{-2.5ex}
     \caption{Visualization of OSKDet detections. (a)(b)(c): DOTA  (d): HRSC2016  (e): UCAS-AOD  (f): ICDAR}
  \label{fig10}
  \end{figure*}

\vspace{-1ex}
\begin{figure}[htbp]
\begin{center}
  \subfigure[standard CNN]{ 
    \label{fig11_1} 
    \includegraphics[width=1.04in]{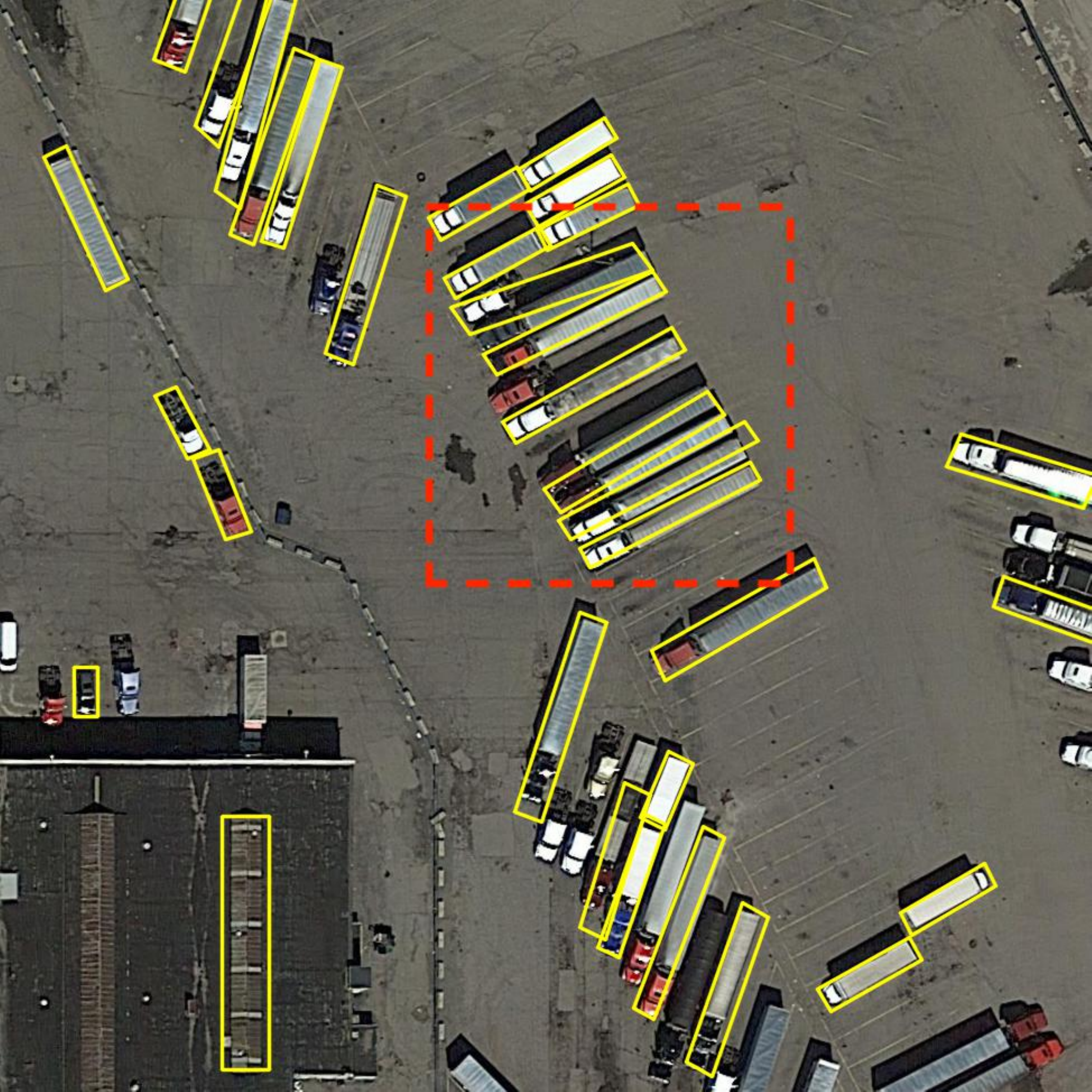} 
  }\hspace{-0.95ex}
  \subfigure[DCN]{ 
    \label{DCN} 
    \includegraphics[width=1.04in]{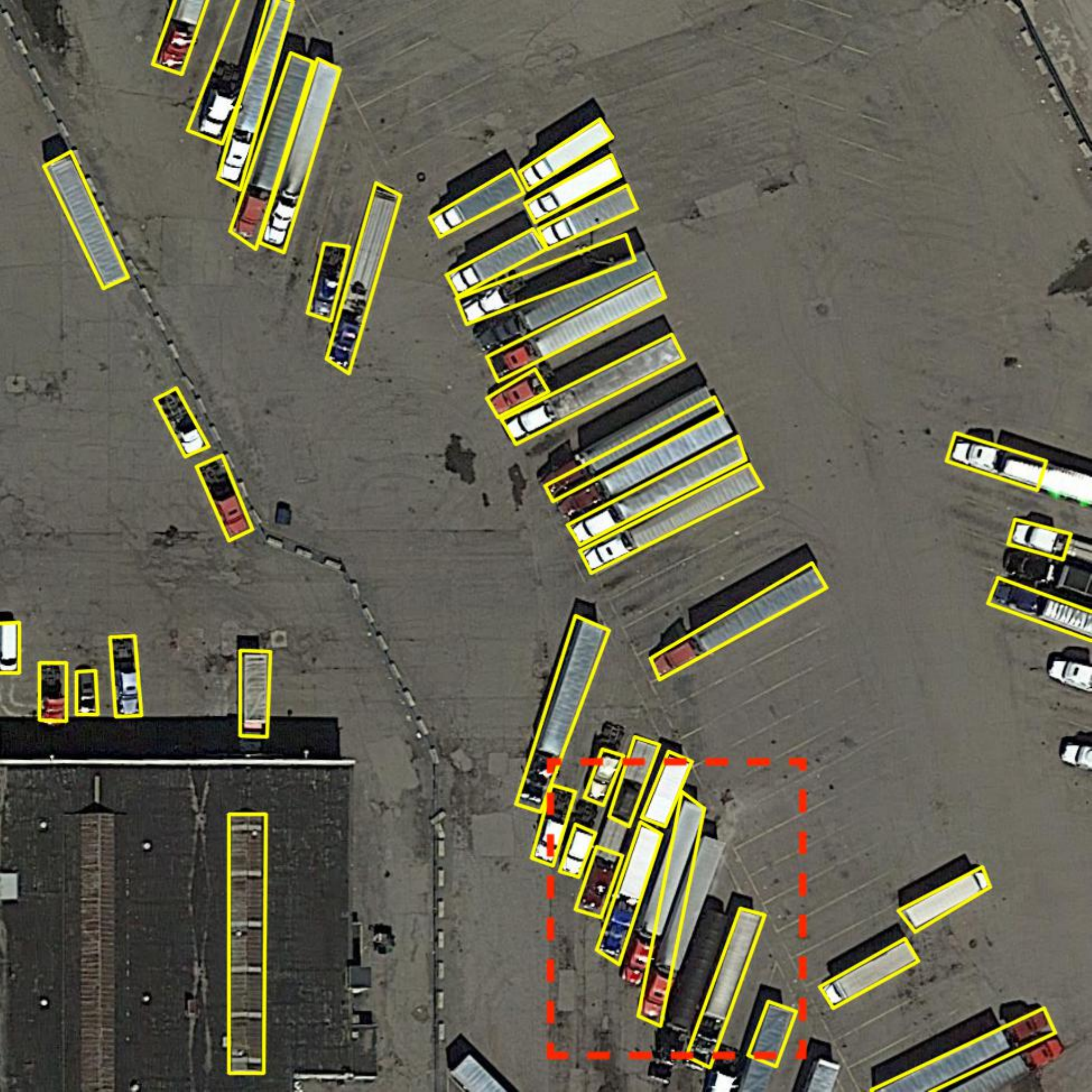} 
  }\hspace{-0.95ex}
  \subfigure[RCN]{ 
    \label{RCN} 
    \includegraphics[width=1.04in]{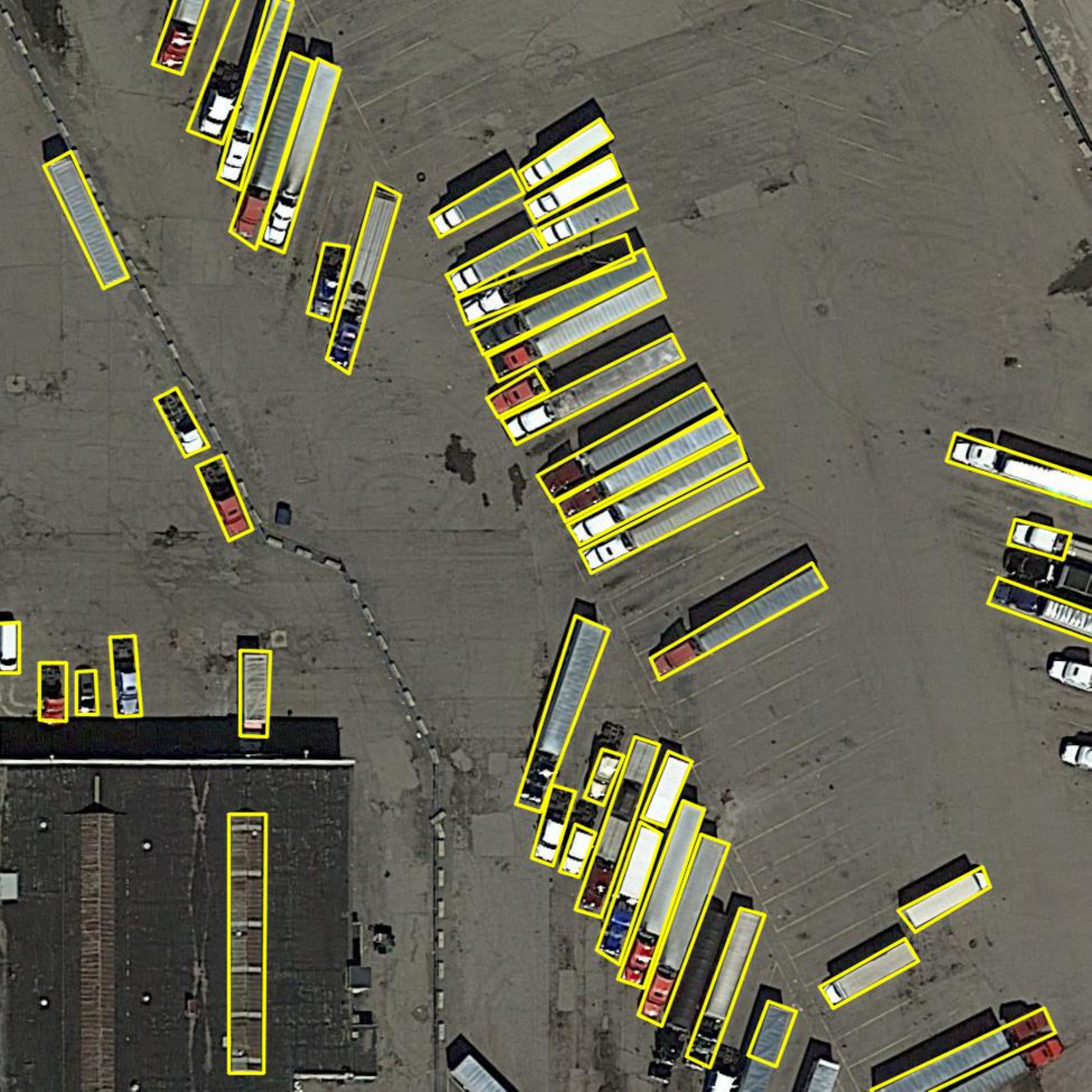} 
  }
\end{center}
\vspace{-2ex}
  \caption{Different CNN modules detections. The red dashed box in (a)(b) is marked as false detection, where the network cannot accurately detect whether keypoints belong to the same target.}

\label{fig11} 
\end{figure}


\textbf{Keypoint reorder and multi-path feature fusion.} We compare the original annotations (unsorted), 0$^{\circ}$, 44$^{\circ}$ as the keypoint sorting interface, modulated loss\cite{rsdet} optimization method, 44$^{\circ}$ sorting and modulated loss (after training 2 epochs, use the modulated loss). Experiment results in Tab \ref{table9} show that our keypoint reorder method performs better than other methods, which improves AP by 0.99\% compared to the baseline. We also study the precision of each angle interval in validation set. The precision near 44$^{\circ}$ is relatively low, but it maintains high accuracy in all other angle range, which proves the effectiveness of our algorithm for eliminating most of the samples confusion. The proposed feature fusion module also has a significant effect on localization precision by improving 0.65\% AP, which further proves the significance of spatial information.

\vspace{-1.5ex}
\begin{table}[htbp]
\centering
\begin{scriptsize}
\begin{tabular}{l|llll|l}
\toprule
Method       &KR-0    &KR-44          &MFF    & ML   & AP.5 \\ \hline
           &              &                &         &           & 75.11        \\
           &  \checkmark   &          &       &        & 75.49(+0.38)        \\
           &          &   &          &          \checkmark       & 75.96(+0.85)     \\
OSKDet     &    &\checkmark &         &               & 76.10(+0.99)     \\
           &   & \checkmark &\checkmark                  &     & 76.75(+1.64)     \\
           &   & \checkmark  &\checkmark     & \checkmark       & \textbf{76.79(+1.68)}     \\ \bottomrule
\end{tabular}
\end{scriptsize}
\vspace{0.8ex}
\caption{Comparison of different keypoint sort methods. ML, KR-0, KR-44 and MFF mean modulated loss, 0$^{\circ}$ sorting interface, 44$^{\circ}$ sorting interface and multi-path feature fusion respectively.}
\label{table9}
\end{table}




\subsection{Text detection}
The text dataset has a greater challenge due to the large number of irregular quadrilaterals and large aspect ratio long text. Tab \ref{table_icdar} shows our detection results on ICDAR2015\cite{ICDAR2015} and ICDAR2017MLT\cite{icdar2017}. OSKDet achieves 91.34\% F-measure on IC15 and 80.29\% F-measure on IC17, which are comparable to the state-of-the-art algorithms. Specifically, On IC15, OSKDet surpasses most of the state-of-the-art methods. Although there are still some gaps with \cite{textfusenet}, OSKDet does not need complex labeling and multi-task calculations in \cite{textfusenet}, which could avoid tedious preprocess and achieves a relatively fast inference speed. Our model could easily extend to other datasets. On IC17, without other techniques designed for text targets, OSKDet surpasses the state-of-the-art method by 0.18\% F-measure, which proves the potential of OSKDet in OCR. 

\vspace{-1ex}

\begin{table}[htbp]
\begin{scriptsize}
\begin{center}
\begin{tabular}{l|l|lll|l}
    \toprule
Task                     & Method                             & Recall    &Precision   &F-measure      & FPS\\ \hline
\multirow{6}{*}{ICDAR2015}        & EAST\cite{EAST}                    &73.50      &83.60       &78.20  & 6.52  \\
                         & PixelLink\cite{deng2018pixellink}  &82.00      &85.50       &83.70   &  7.30   \\
                         & PSENet\cite{wang2019shape}         &85.22      &89.30       &87.21   & 2.33   \\
                          & FOTS\cite{FOTS}                    &85.17      &91.00       &87.99 & 7.50 \\
                         & Textfusenet\cite{textfusenet}      &89.70      &94.70       &\textbf{92.10}& 4.10\\
                         & OSKDet(ours)                     &88.35      &94.56       &91.34 & 6.31    \\ \hline
\multirow{4}{*}{ICDAR2017}                          & PSENet\cite{wang2019shape}         &68.35      &76.97       &72.40 &-  \\
                         & SBD\cite{sbd}                      &70.10      &83.60       &76.30      &- \\
                         & PMTD\cite{pmtd}                    &76.25      &84.42       &80.13   &-  \\
                         & OSKDet(ours)                     &75.69      &85.53       &\textbf{80.31}  & 6.01 \\ \bottomrule
\end{tabular}
\end{center}
\end{scriptsize}
\vspace{-1ex}
\centering
\caption{Comparison of different methods on ICDAR}
\label{table_icdar}
\end{table}


\vspace{-3.5ex}



\section{Conclusion}

\vspace{-0.5ex}




This paper proposes a rotated object detection model OSKDet. By the proposed orientation-sensitive heatmap and rotation-aware deformable convolution, OSKDet can extract spatial feature effectively and obtain high quality detection results. The proposed keypoint sorting methods and feature fusion module eliminate the confusion of keypoint order greatly. Experimental results on several public datasets show the state-of-the-art performance of OSKDet.

{\small
\bibliographystyle{ieee_fullname}
\bibliography{main}
}

\end{document}